\documentclass[10pt,twocolumn,letterpaper]{article}

\usepackage[pagenumbers]{cvpr}               %
\usepackage{amsmath}
\usepackage{amssymb}
\usepackage{graphicx}

\usepackage[usestackEOL]{stackengine}

\newif\ifsiggraph
\siggraphfalse

\newcommand{\ourmlp}{Concept-Mod}

\ifsiggraph
\newcommand{\eg}{e.g.,\xspace}
\newcommand{\ie}{i.e.,\xspace}
\else
\fi

\usepackage{xspace}

\usepackage{soul}

\definecolor{cvprblue}{rgb}{0.21,0.49,0.74}
\usepackage[pagebackref,breaklinks,colorlinks,allcolors=cvprblue]{hyperref}

\title{TokenVerse: Versatile Multi-concept Personalization in Token Modulation Space}
\vspace{-0.5cm}

\author{
        Daniel Garibi$^{1,2,*}$ \hspace{5mm}
        Shahar Yadin$^{1,3,*}$ \hspace{5mm}
        Roni Paiss$^1$ \hspace{5mm}
        Omer Tov$^1$ \hspace{5mm}
        Shiran Zada$^1$
        \\[3pt]
        Ariel Ephrat$^1$ \hspace{5mm}
        Tomer Michaeli$^{1,3}$ \hspace{5mm}
        Inbar Mosseri$^1$ \hspace{5mm}
        Tali Dekel$^{1,4}$ 
        \\ [4pt]
        $^1$Google DeepMind \hspace{10mm} $^2$Tel Aviv University \hspace{10mm} $^3$Technion \hspace{10mm} $^4$Weizmann Institute 
        \\
        \small\url{https://token-verse.github.io/}
         \\[-30pt]
         }

\begin{document}

\ifsiggraph
    \begin{teaserfigure}
        \centering
        \setlength{\tabcolsep}{1pt}
        \includegraphics[width=\textwidth]{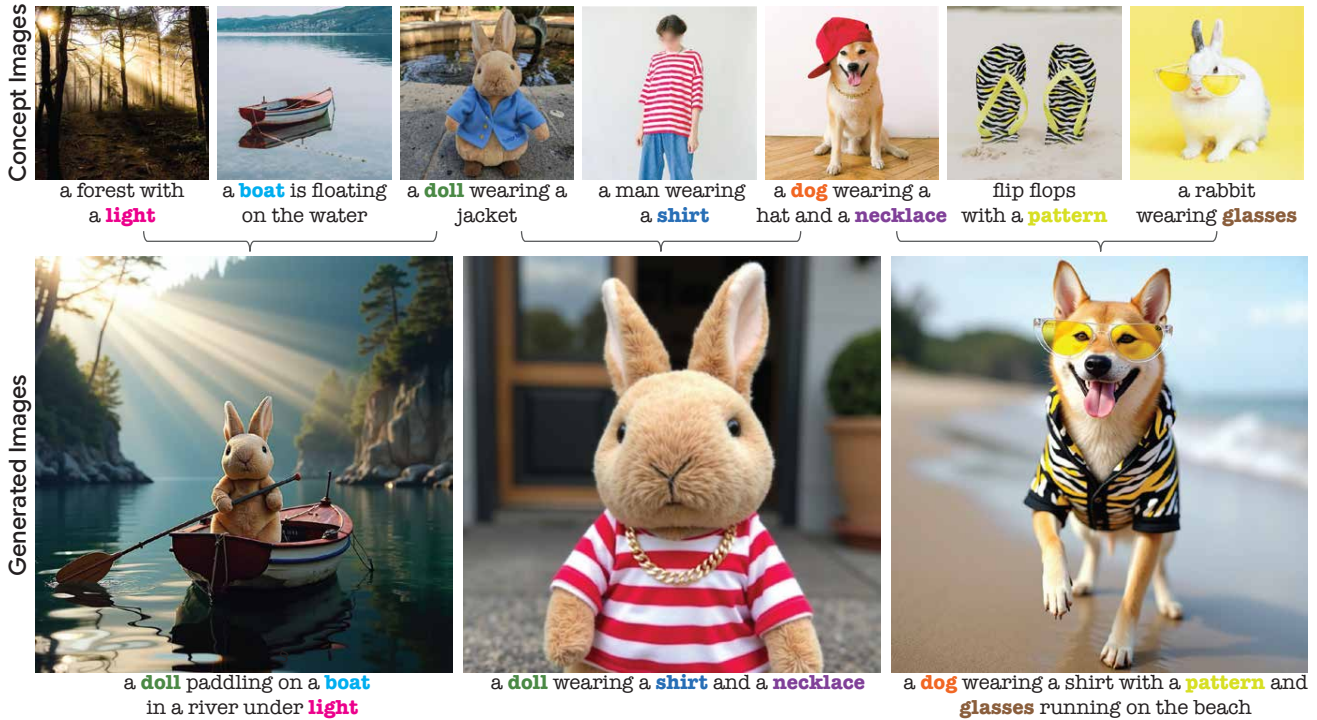}
        \vspace{-0.7cm}
        \caption{TokenVerse extracts distinct complex visual concepts from a set of \emph{concept images} (top), and allows users to generate images that depict these concepts in novel versatile compositions (bottom row). Our framework \emph{independently} processes each concept image, and learns to disentangle its concepts based solely on an accompanying caption, without any additional supervision or masks. This is achieved by learning a personalized representation for each token in the source caption.  Our personalized text tokens, extracted from multiple images, are then flexibly incorporated into new text prompts (colored words) to generate novel creative images.}
        \label{fig:teaser}
    \end{teaserfigure}
\else
    \twocolumn[{%
        \renewcommand\twocolumn[2][]{#1}%
        \maketitle
        \begin{center}
            \centering \centering
            \includegraphics[width=\textwidth]{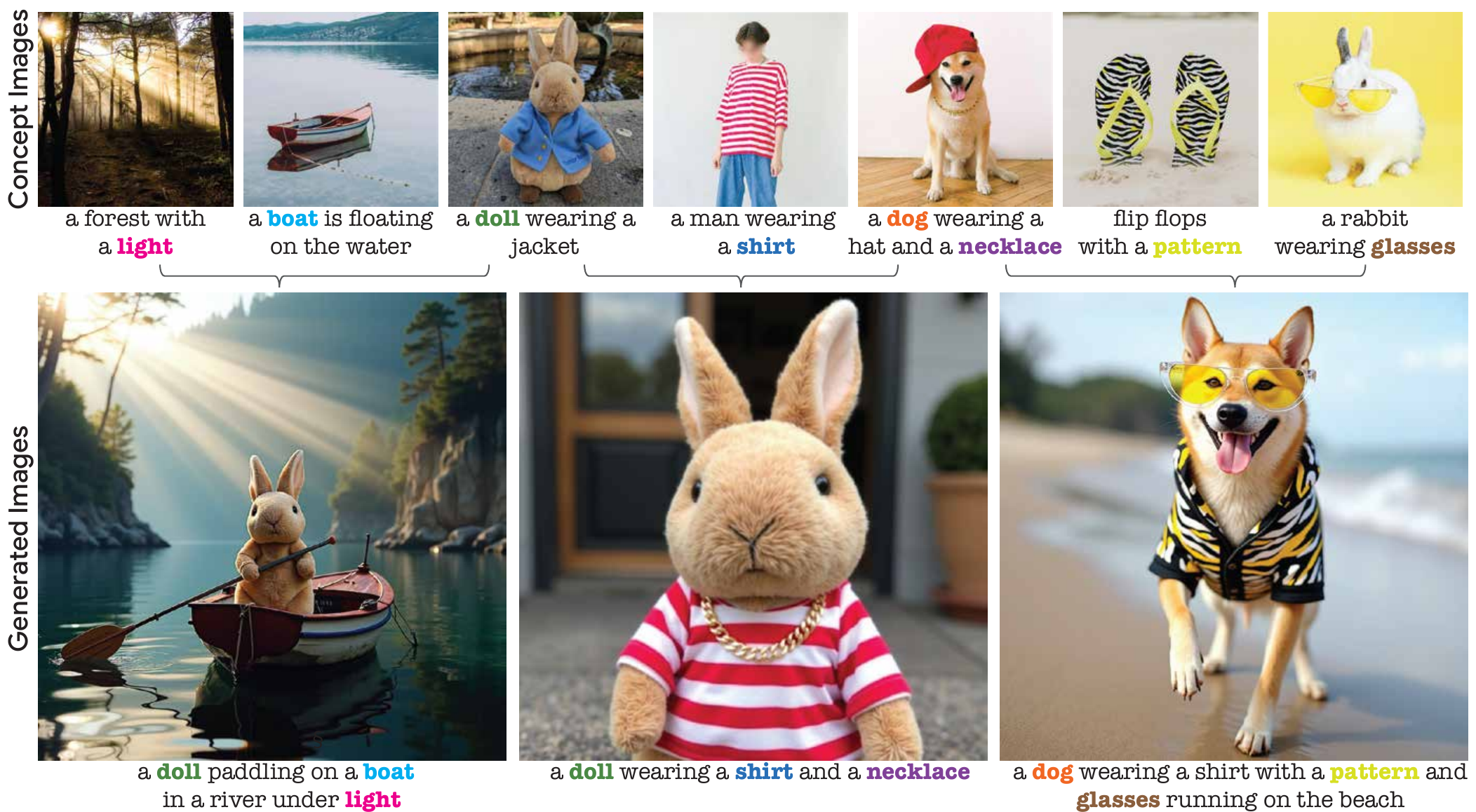}
            \vspace{-0.5cm}
            \captionof{figure}{TokenVerse extracts distinct complex visual concepts from a set of \emph{concept images} (top), and allows users to generate images that depict these concepts in novel versatile compositions (bottom row). Our framework \emph{independently} processes each concept image, and learns to disentangle its concepts based solely on an accompanying caption, without any additional supervision or masks. This is achieved by learning a personalized representation for each token in the source caption.  Our personalized text tokens, extracted from multiple images, are then flexibly incorporated into new text prompts (colored words) to generate novel creative images. }
        \label{fig:teaser}
        \end{center}
    }] 
\fi

\def\thefootnote{*}\footnotetext{Equal contribution, work was done while interns at Google DeepMind}
\begin{abstract}
We present TokenVerse -- a method for multi-concept personalization, leveraging a pre-trained text-to-image diffusion model. Our framework can disentangle complex visual elements and attributes from as little as a single image, while enabling seamless plug-and-play generation of combinations of concepts extracted from multiple images. As opposed to existing works, %
TokenVerse can handle multiple images with multiple concepts each, and supports a wide-range of concepts, including objects, accessories, materials, pose, and lighting.  Our work exploits a DiT-based text-to-image model, in which the input text affects the generation through both attention and modulation (shift and scale). We observe that the modulation space is semantic and enables localized control over complex concepts. Building on this insight, we devise an optimization-based framework that takes as input an image and a text description, and finds for each word  a  distinct direction in the modulation space. These directions can then be used to generate new images that combine the learned concepts in a desired configuration. We demonstrate the effectiveness of TokenVerse in challenging personalization settings, and showcase its advantages over existing methods. %
\end{abstract}

\section{Introduction}

The visual world is a rich tapestry of visual concepts, encompassing diverse objects, poses, lighting conditions, materials, and textures, often intricately combined in complex ways. Text-to-image models have shown a remarkable ability to learn and represent this complexity, generating images that not only capture individual concepts but also seamlessly integrate them into diverse and coherent settings. In this work, our goal is to gain user-control over the unique properties of the generated concepts. %
Specifically, our approach seeks to: $(i)$ disentangle and learn the distinctive attributes of various visual elements within as little as a single image, and $(ii)$ enable the generation of new images that flexibly combine different subsets of concepts extracted from multiple images (Fig.~\ref{fig:teaser}). 
Such versatile control over a diverse array of concepts is a pivotal ingredient in  various real-world content creation tasks, including storytelling and personalized content creation.
Yet, existing approaches focus on personalization only for objects or styles, often require segmentation masks or bounding boxes, and can either support a single concept per image or multiple concepts within a single image.

Broadly speaking, there are two main approaches for personalized content creation. 
The first is fine-tuning a text-to-image generative model \cite{ruiz2023dreamboothfinetuningtexttoimage}, teaching it to associate some unique text token with a new concept. This approach is restricted in its ability to seamlessly compose multiple concepts, as it requires the combination of model weights, each specialized for a different concept. The second approach keeps the model fixed and optimizes the input text embedding associated with a word that describes a concept \cite{gal2022imageworthwordpersonalizing}.  
However, this approach is often not expressive enough to fully capture the nuances of each concept. Furthermore, both approaches are primarily designed for capturing a single concept, and struggle to disentangle multiple concepts encapsulated in a single image. While various methods have been developed to generalize these approaches (\eg by constraining different concepts to represent distinct image regions), existing methods are still restrictive in either the type or breadth of concepts they can handle. In particular, they struggle to disentangle non-object concepts like pose, materials, and lighting conditions. %

Our objective is to enable versatile and flexible concept personalization, where diverse concepts are extracted from single images, and can be combined in different configurations. Our framework  builds on a pre-trained text-to-image Diffusion Transformer (DiT) model, in which the input text is processed in two paths: $(i)$ through transformer blocks where the text tokens are jointly processed with the image tokens, and $(ii)$ through a modulation path where a global text embedding is mapped to scale and shift coefficients that modulate the channels of the tokens within each transformer block. Inspired by the successful use of the modulation space in GANs for semantic image manipulation, here we explore the use of the modulation space in DiTs for our task. We observe that directions within this space, which we call $\mathcal{M}$, correspond to semantic modifications to the generated image, similarly to GANs. However, these manipulations are often not sufficiently localized for personalized content creation. Yet, we find that modifying the modulation vector for \emph{only a single text token} can lead to semantic modifications only to the concept associated with this token. We denote the space of per-token modulations by $\mathcal{M}^+$.

We take advantage of the space $\mathcal{M}^+$ for unsupervised disentangled personalization and composition of visual concepts. We show that given an image and its caption, optimizing a modulation vector for each text token is sufficient to personalize the visual element it describes.
The disentanglement of the visual concepts in the image is naturally facilitated by the model's inherent association between text tokens and their corresponding image parts \cite{hertz2022prompttoprompt}. While the modulation vectors for all text tokens are optimized jointly, we find that each optimized vector personalizes the visual element tied to the text token it modulates. Thus, multiple learned elements can be generated together in new settings by simply describing their composition in the caption and plugging the learned vectors for their text tokens.

We show that our approach is expressive enough to represent complex concepts without requiring adjustment to the model's weights, thus preserving its prior. In addition, as the visual elements are defined by semantic text tokens rather than visual cues (\eg segmentation masks), our method supports the personalization of overlapping objects, as well as of non-object concepts (\eg poses, lighting and materials) simply by describing them in the caption during optimization  (Fig.~\ref{fig:teaser}). Lastly, this approach is highly modular, allowing concepts extracted from multiple images to be seamlessly combined.

To summarize, our key contributions are as follows:
\begin{itemize}
    \item We present TokenVerse -- the first method that enables disentangled multi-concept personalization from multiple images and the plug-and-play composition of new images that depict the learned concepts. 
    \item Our method can personalize semantic concepts beyond objects, such as lighting conditions, materials and poses. 
    \item We explore the role of text token modulation in the generation process in DiTs, and demonstrate its effectiveness as a localized and semantically-meaningful space.  
    \item We demonstrate the applicability of TokenVerse for personalized content creation and storytelling.
\end{itemize}

\section{Related work}
\label{sec:related}

\paragraph{Diffusion transformers.}

While diffusion models initially relied on UNet-based architectures \cite{rombach2022highresolution, podell2023sdxl, ramesh2022hierarchical, saharia2022photorealistic}, most recent state-of-the-art text-to-image models use diffusion transformers (DiTs) as their backbone \cite{esser2024scalingrectifiedflowtransformers, flux}.
Unlike in UNets, where the text condition is queried via cross-attention layers, in DiTs the text tokens are processed by the transformer alongside the image tokens.
Joint attention layers facilitate bidirectional interaction between image and text tokens, enabling seamless integration of textual and visual information throughout the generation process.

\paragraph{Personalization methods.}

Personalization methods extend the distribution of a pretrained model to include a specific concept provided by a set of images.
Most approaches achieve this either by learning specialized text embeddings to represent the concept \cite{gal2022imageworthwordpersonalizing, alaluf2023neuralspacetimerepresentationtexttoimage, voynov2023pextendedtextualconditioning} or by fine-tuning layers of the model itself \cite{ruiz2023dreamboothfinetuningtexttoimage, shah2023ziplorasubjectstyleeffectively, frenkel2024implicitstylecontentseparationusing}.
These methods learn a single concept from one or several images, and can then incorporate the learned concept in new images generated according to text prompts.
In contrast, we focus on multi-concept disentangled personalization and composition, \ie extracting several concepts from a single image, and composing them with concepts learned from other images.

\ifsiggraph

\begin{figure}
    \centering
    \includegraphics[width=\linewidth]{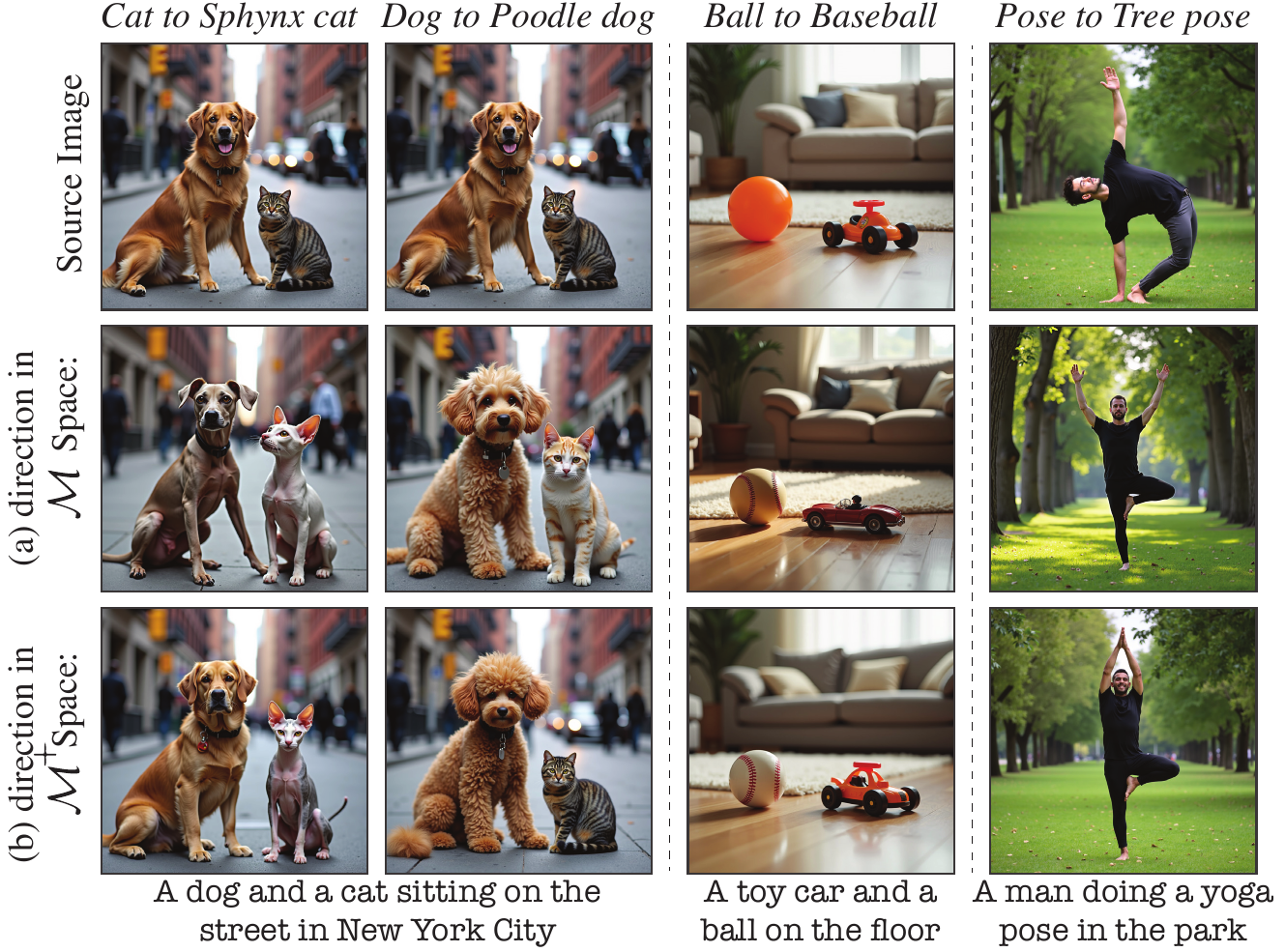}
    \ifsiggraph
        \vspace{-0.75cm}
    ֿ\else
    \fi
    \caption{\textbf{Directions in the global modulation space ($\mathcal{M}$) and our per-token modulation space ($\mathcal{M}^+$). }
    Given a generated image (top row), we modify it using text-driven directions in both $\mathcal{M}$ and $\mathcal{M}^+$ spaces.
    (a) Adding a direction to the vector that is used to modulate all the text and image tokens (\ie a direction in the space $\mathcal{M}$) can be used to effectively modify desired concepts in the generated image. Yet, this often results in non-local changes that also affect other concepts in the generated image. (b)~Adding a direction only to the modulation vector of a specific text token, like ``dog'' or ``ball'' (\ie  a direction in the space $\mathcal{M}^+$) leads to a localized modification that mostly affects the concept of interest.} 
    \label{fig:editing_directions}

    \ifsiggraph
        \vspace{-0.75cm}
    ֿ\else
    \fi
\end{figure}

ֿ\else
    
\fi

\paragraph{Disentangled multi-concept personalization}
Several recent works aimed to learn multiple concepts from a single or few images, enabling these concepts to be reused in novel compositions. Break-a-Scene~\cite{Avrahami_2023} employs Dreambooth finetuning~\cite{ruiz2023dreamboothfinetuningtexttoimage} while relying on user-provided spatial masks to isolate the concepts in the image. The dependence on spatial masks poses a significant limitation, as does not allow to disentangle \eg the appearance of an object from its pose. %
Other methods, like Inspiration Tree~\cite{vinker2023conceptdecompositionvisualexploration} and ConceptExpress~\cite{hao2024conceptexpressharnessingdiffusionmodels}, extract multiple concepts by jointly learning several tokens from a single image, each representing a different concept.
However, the disentanglement achieved by these methods is unpredictable, offering no control over which aspects of the image are separated into individual concepts. This limits the flexibility and usability of the approach, making it less effective for generating complex and diverse compositions. Unlike these approaches, our method provides control over the visual elements that should be personalized without requiring any visual cues.

\paragraph{Composition of concepts}
Complementary to the task of multi-concept personalization, is that of combining concepts from different images into a single generated image.
Methods for concept composition typically utilize LoRA \cite{hu2021loralowrankadaptationlarge} to learn new concepts (usually one concept per image) and propose techniques for merging multiple LoRAs to operate simultaneously.
Some methods, like LoRA Composer~\cite{yang2024loracomposerleveraginglowrankadaptation} and OMG~\cite{kong2024omgocclusionfriendlypersonalizedmulticoncept}, 
achieve this by incorporating spatial conditioning, such as masks, and assigning localized prompts to specific regions of the image. %
However, this limits the ability to compose concepts that naturally overlap, such as a necklace worn by a man. 
Other approaches~\cite{gu2023mixofshowdecentralizedlowrankadaptation, po2023orthogonal, shah2023ziplorasubjectstyleeffectively} fuse multiple LoRAs, allowing them to work cohesively. 
However, these methods are either restricted to composing style and content or require a joint optimization process for all source images, limiting the number of distinct images that can be effectively combined. Contrary to these methods, our approach supports plug-and-play composition of a much larger number of personalized objects, without any user-provided spatial conditioning.

\ifsiggraph
ֿ\else

\begin{figure}
    \centering
    \includegraphics[width=\linewidth]{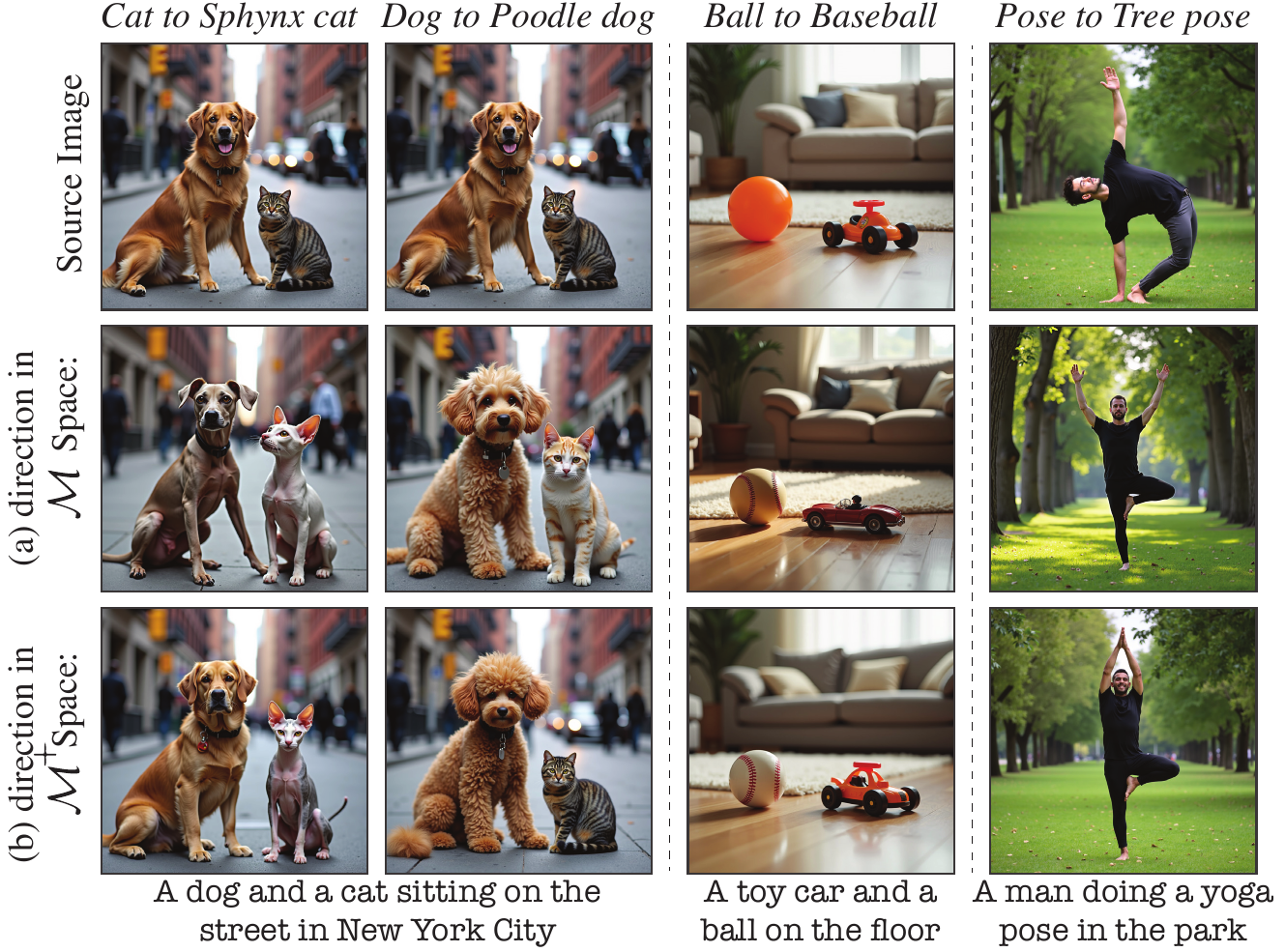}
    \ifsiggraph
        \vspace{-0.75cm}
    ֿ\else
    \fi
    \caption{\textbf{Directions in the global modulation space ($\mathcal{M}$) and our per-token modulation space ($\mathcal{M}^+$). }
    Given a generated image (top row), we modify it using text-driven directions in both $\mathcal{M}$ and $\mathcal{M}^+$ spaces.
    (a) Adding a direction to the vector that is used to modulate all the text and image tokens (\ie a direction in the space $\mathcal{M}$) can be used to effectively modify desired concepts in the generated image. Yet, this often results in non-local changes that also affect other concepts in the generated image. (b)~Adding a direction only to the modulation vector of a specific text token, like ``dog'' or ``ball'' (\ie  a direction in the space $\mathcal{M}^+$) leads to a localized modification that mostly affects the concept of interest.} 
    \label{fig:editing_directions}

    \ifsiggraph
        \vspace{-0.75cm}
    ֿ\else
    \fi
\end{figure}

\fi

\section{Preliminaries: Diffusion transformers}
\begin{figure*} [t]
    \centering
    \includegraphics[width=\linewidth]{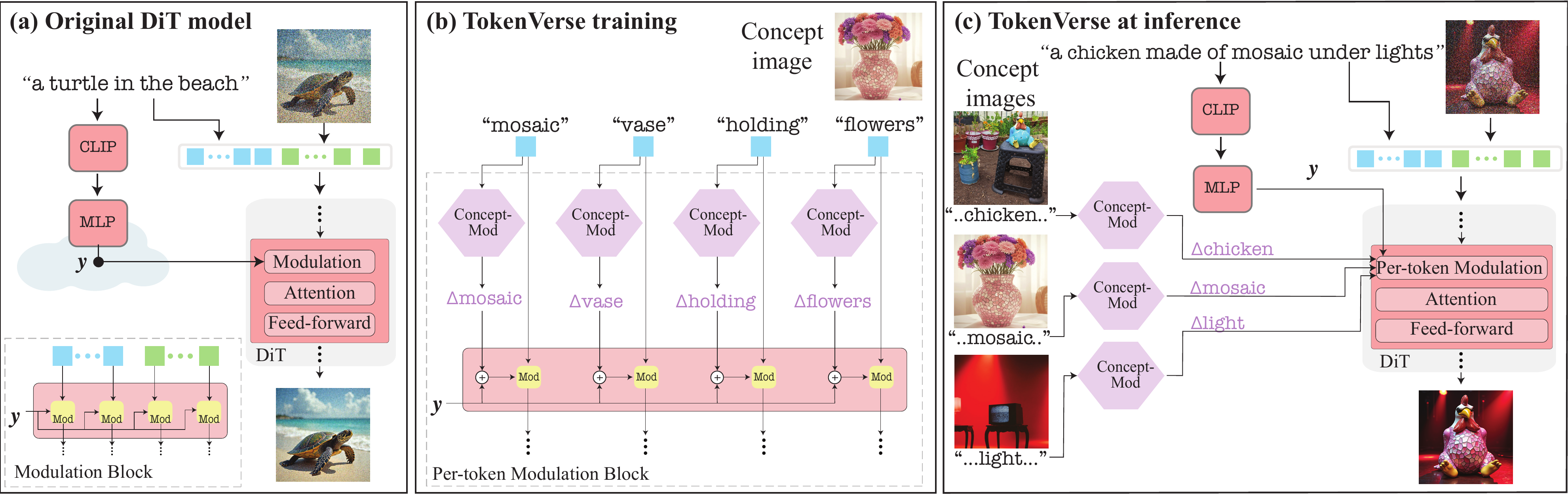}
    \ifsiggraph
    \vspace{-0.75cm}
    \else
    \fi
    
    \caption{\textbf{TokenVerse overview.}
    (a) A pre-trained text-to-image DiT model processes both image and text tokens via a series of DiT blocks. Each block consists of modulation, attention and feed-forward modules.  We focus on the modulation block, in which the tokens are modulated via a vector $y$, which is derived from a  pooled text embedding. (b) Given a concept image and its corresponding caption, TokenVerse learns a personalized modulation vector offset $\Delta$ for each text token. These offsets represent personalized directions in the modulation space and are learned using a simple reconstruction objective. (c) At inference, the pre-learned direction vectors are used to modulate the text tokens, enabling the injection of personalized concepts into the generated images.
    }
    \label{fig:method_overview}
    \ifsiggraph
    \vspace{-0.375cm}
    \else
    \fi
\end{figure*}

Diffusion and flow-based models generate images through an iterative process that starts with a sample of white Gaussian noise and gradually evolves into a valid sample. In each step, the current noisy image is passed through a neural network, which is used to obtain a slightly less noisy version of the image. As illustrated in Fig.~\ref{fig:method_overview}(a), in text-to-image diffusion transformers (DiTs), this neural network is a transformer that jointly processes the text and image tokens. Each DiT block contains an attention layer, a feed-forward MLP, and a modulation mechanism that serves to incorporate conditioning signals, as we explain next.

\paragraph{The Modulation mechanism in DiTs.}

modulation refers to modifying the activations of a neural network on a per-channel basis, where each channel is multiplied by a single scale factor and shifted by a single bias scalar. In the context of generative models, the modulation mechanism has gained significant popularity following its use in StyleGAN \cite{karras2017progressive,karras2019style,karras2020analyzing,Karras2020ada,karras2021alias}. There, small modifications to the modulation parameters have been shown to lead to smooth and semantically meaningful perturbations to the generated image, a property that has been exploited in a wide range of works on image editing and manipulation \cite{abdal2019image2stylegan,abdal2020image2stylegan++,abdal2020styleflow,harkonen2020ganspace,patashnik2021styleclip,richardson2020encoding,tov2021designing,alaluf2021hyperstyle,alaluf2021restyle,roich2022pivotal,zhu2020domain}. In modern text-to-image DiTs, the modulation mechanism is used for incorporating conditioning signals, such as the diffusion timestep and a compact representation of the text prompt. In particular, Stable Diffusion~3 and Flux \cite{esser2024scalingrectifiedflowtransformers, flux} process the diffusion timestep $t$ and a pooled embedding of the text prompt~$p$ (CLIP's class token) by an MLP, which outputs a vector%
\begin{equation}
\label{eq:y_eq}
y = \text{MLP} (t, \text{CLIP}(p)).
\end{equation}
This vector is then further processed and split into per-channel scale and shift parameters, which are used to modulate the text and image tokens within each block of the diffusion model. Importantly, the same scale and shift parameters are used for all tokens. %

\section{The \texorpdfstring{$\mathcal{M}^+$}{} space}

Inspired by the use of the modulation space in StyleGAN for editing, we now explore the space of modulation vectors in DiTs, which we coin $\mathcal{M}$. We start by demonstrating the effect of modifying the modulation vector using simple manipulations on the text embedding injected to the modulation path (keeping the text at the input of the transformer fixed). We show this allows achieving semantically rich modifications, but often changes the generated image in a non-localized manner. We then introduce our proposed framework, which modifies only the modulation vector that affects specific text tokens. We coin the space of all such modifications the $\mathcal{M}^+$ space.

\paragraph{The modulation space $\mathcal{M}$.}

A naive approach to obtaining a direction in the space of modulation vectors is by using text prompts with and without a specific attribute,
\begin{equation}
\label{eq:delta}
\Delta_{\text{attribute}} =  \text{MLP}(t, e_{\text{attribute}}) -  \text{MLP}(t, e_{\text{neutral}}),
\end{equation}
where $e_{\text{neutral}}$ is the pooled embedding of the text prompt used to generate the image and $e_{\text{attribute}}$ is the pooled embedding of the same prompt but with some attribute added to the object of interest (\eg ``Poodle dog'' instead of ``dog''). Having obtained such a direction, we can add it to the modulation vector $y$ (Eq.~\eqref{eq:y_eq}) with some scale factor $w$, to obtain an updated modulation vector, $y + w \Delta_{\text{attribute}}$. 

Figure~\ref{fig:editing_directions} (a) illustrates the effect that this has on the generated image. As can be seen, the object of interest is indeed modified into possessing the desired attribute. However, the modification is not localized. Namely, each direction modifies also unrelated attributes in the image (\eg~changing both the dog and the cat when only one of them is specified in the direction).

\paragraph{The Per-token modulation space $\mathcal{M}^+$.}
To overcome the lack of localization, here we propose to modulate individual text tokens differently. We coin the space corresponding to all per-text-token modulation vectors $\mathcal{M}^+$. 
Specifically, rather than using the same vector $y$ to modulate all tokens, we propose modifying the modulation vector only for the text token corresponding to the concept we wish to affect. %
Modifying the modulation vector for a single text token affects the corresponding image tokens through the joint attention layer, and this turns out to translate into more localized effects in the areas of the image associated with that token.

Figure~\ref{fig:editing_directions}(b) illustrates the effect of such directions. %
For changes to the dog we apply the modulation vector $y + w \Delta_{\text{Poodle}}$ (Eq.~\eqref{eq:delta}) only to the ``dog'' text token while using the unmodified $y$ for the rest of the tokens. Similarly, for affecting the cat we apply the modulation vector $y + w \Delta_{\text{Sphynx}}$ only to the ``cat'' text token. As can be seen, these  directions lead to highly localized changes, mostly affecting the objects corresponding to the manipulated tokens. This approach is not limited to objects and can also be applied to abstract concepts like pose. For example, in the case of a man doing yoga, the pose changes when modifying the modulation vector in the direction $ \Delta_{\text{Tree Pose}}$. In this case as well, modifying the modulation vector only for the word ``pose'' better preserves the background of the image compared to modifying all tokens.

\section{Disentangled concept learning}
\label{sec:method}

Having observed that the space $\mathcal{M}^+$ enables localized semantic manipulations, we would now like to find customized directions within this space. Specifically, given an example image depicting several desired concepts (a ``concept image'') and a caption describing it, our goal is to learn disentangled representations for each of the visual concepts mentioned in the caption. Importantly, we aim to achieve this in an unsupervised manner, without relying on object masks. For example, for the caption ``a person dancing at dawn'', we want to associate the words ``person'', ``dancing'', and ``dawn'' with  directions in $\mathcal{M}^+$ that capture the identity, pose and lighting in the image, respectively.  
Once we extract these directions from the concept image, we can use them to generate new images with any subset of the learned concepts, simply by adding the directions to the appropriate text tokens. Importantly, our approach is modular in the sense that different directions can be extracted separately from different concept images, without the need for joint training. %

Our objective is to determine a set of directions $\smash{\{\Delta_i\}_{i=1}^{\text{len}(p)}}$, each corresponding to a different token in the prompt $p$, %
which represent the concepts associated with those tokens. As illustrated in Fig.~\ref{fig:method_overview}(b), we do so by training 
\ourmlp{} -- a small MLP that predicts for every token in $p$ a direction in $\mathcal{M}^+$, 
\begin{equation}
(\Delta_1, ..., \Delta_{\text{len}(p)}) = \text\ourmlp(p).
\end{equation}
Our goal is to have each $\Delta _i$ represent the direction in $\mathcal{M}^+$ between the $i$\textsuperscript{th} token  (\eg ``person'', which represents a generic human) and its customized version (\eg the specific person appearing in the image).
We train \ourmlp{} on the concept image and its associated prompt by using the same diffusion objective with which the original text-to-image model was trained. At inference, the learned concepts can be incorporated into newly generated images by adding the learned offsets to the appropriate text tokens (Fig.~\ref{fig:method_overview}(c)).

\paragraph{Per-block optimization.}
We learn the directions in two stages. In the first stage, we aim to learn the coarse aspects of the concepts in the image. This is done by prioritizing the selection of high noise levels in the optimization of the diffusion loss. In the second stage, we refine the directions by focusing more on the lower noise levels. In this stage, we also train an additional MLP that outputs a vector per transformer block. The outputs of this per-block MLP are added to the output of the \ourmlp{} MLP, leading to a per-token per-block direction.

\begin{figure}[t]
    \centering
    \includegraphics[width=\linewidth]{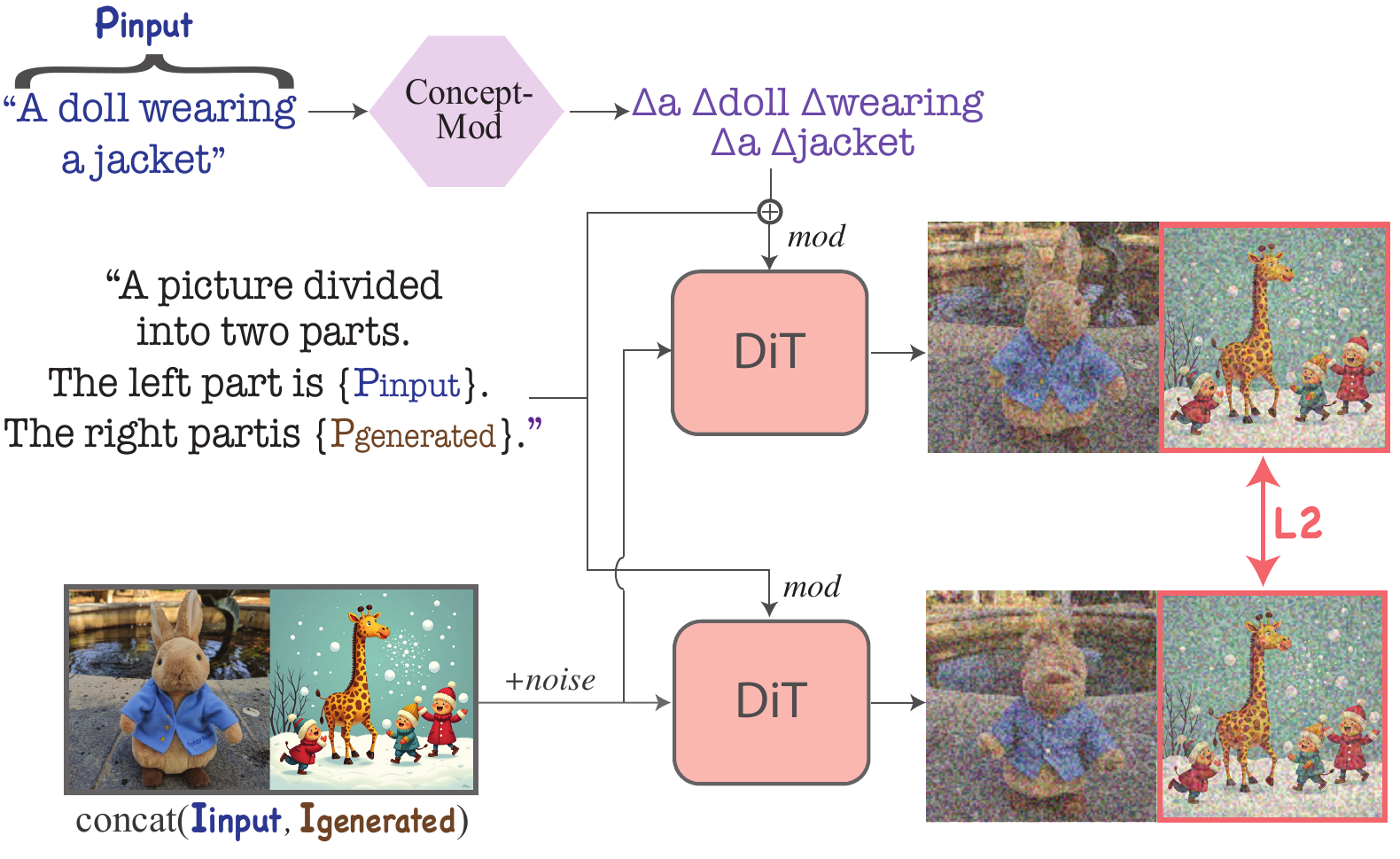}
    \ifsiggraph
    \vspace{-0.75cm}
    \else
    \fi
    
    \caption{\textbf{Concept isolation loss.} When training \ourmlp{} we apply an additional \textit{concept isolation loss} in 50\% of the training steps. This loss encourages learning directions that do not interfere with other images by enforcing that the parts in the image that should not be affected by the directions remain similar.} 
    \label{fig:concept_isolation_loss}
    
    \ifsiggraph
    \vspace{-0.6cm}
    \else
    \fi
\end{figure}

\ifsiggraph
ֿ\else
    \begin{figure*} [t]
    \centering
    \ifsiggraph
        \includegraphics[width=\linewidth]{images/qual14.pdf}
        \vspace{-0.75cm}
    ֿ\else
        \includegraphics[width=\linewidth]{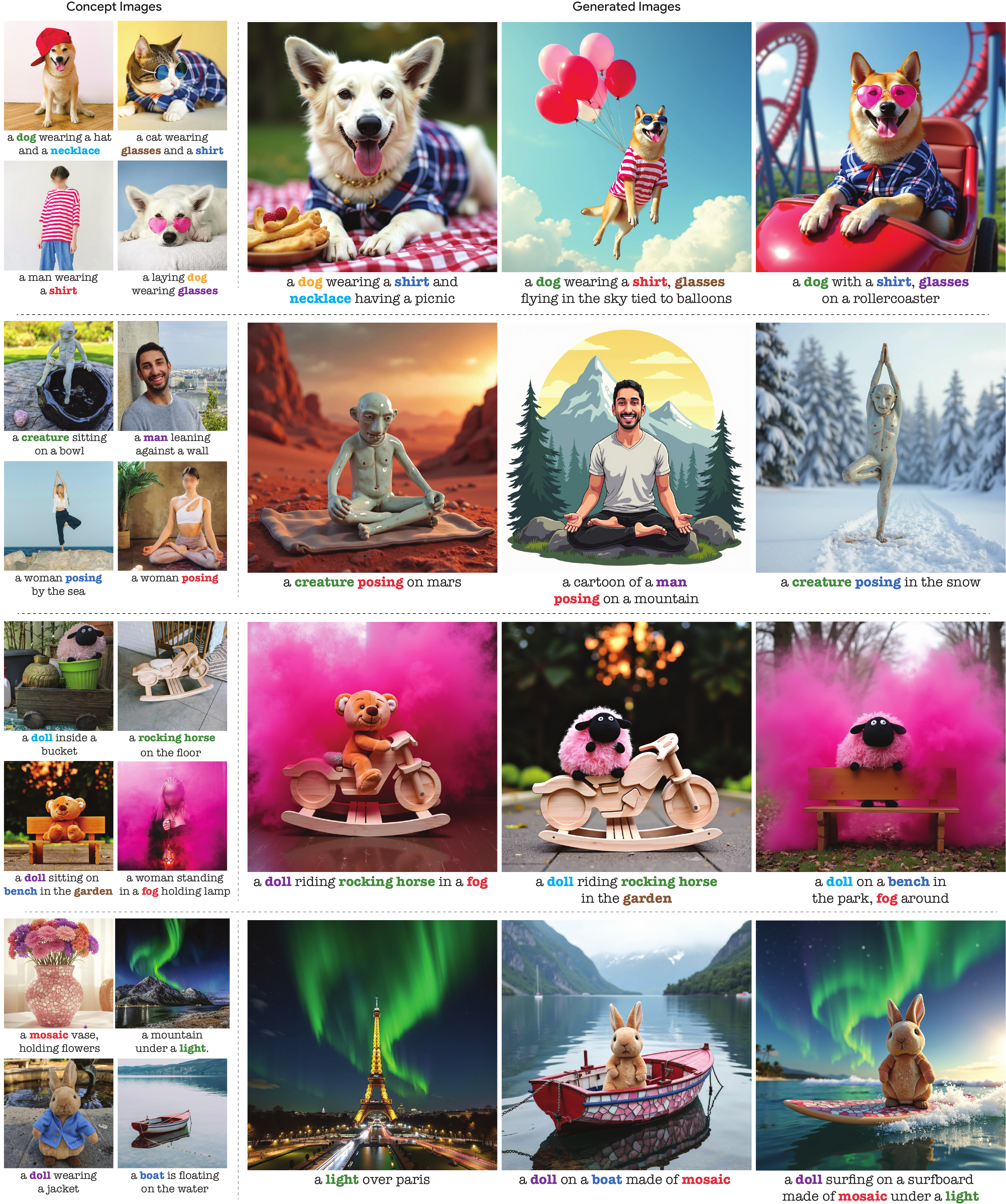}
        \vspace{-0.3cm}
    \fi
    \caption{\textbf{Qualitative results.}
    Each row begins with a bank of four source images, from which our method independently extracts concepts. To the right, three generated images are shown, demonstrating the seamless combination of these concepts into new, coherent outputs.
    }
    \label{fig:qualitative_results}
\end{figure*}

    \begin{figure*} [t]
    \centering
    \ifsiggraph
        \includegraphics[width=\linewidth]{images/qual_res_6-1.pdf}
        \vspace{-0.5cm}
    ֿ\else
        \includegraphics[width=\linewidth]{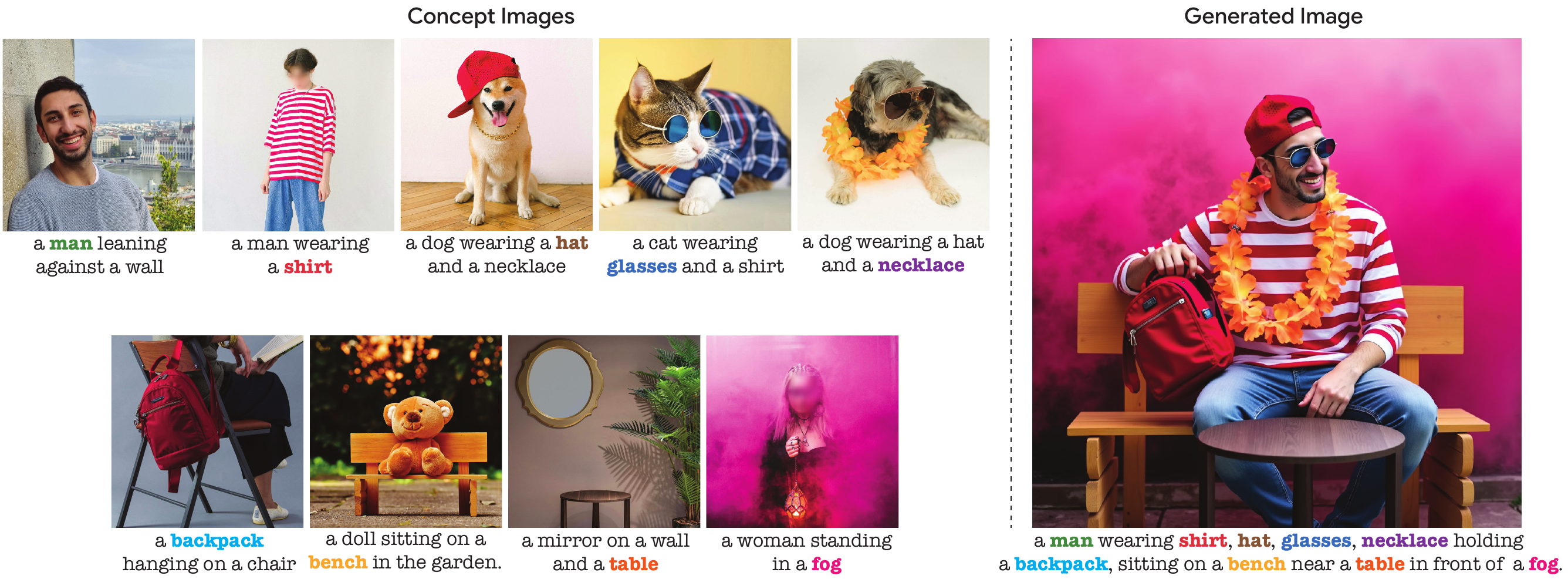}
    \fi
    \caption{\textbf{Extreme multi-concept personalization.} Our method has no technical constraint on the number of concepts that can be combined in an image. As can be seen, TokenVerse can generate images composing a significant number of concepts.
    }
    \label{fig:qualitative_results_6}
    \ifsiggraph
    \vspace{-0.5cm}
    \else
    \fi
\end{figure*}

\fi

\paragraph{Concept isolation loss.}

TokenVerse learns for each token in the caption a separate direction in $\mathcal{M}^+$. As this space is relatively disentangled, directions learned from the same image are usually well isolated. 
However, when combining objects learned from different images, their optimized directions might interfere with each other, and degrade concept fidelity.
To avoid such cases, we incorporate an additional \textit{concept isolation loss} in 50\% of the training iterations. 
This loss is designed to steer the optimization such that the optimized directions do not affect concepts that do not appear in the concept image.
As illustrated in Fig.~\ref{fig:concept_isolation_loss}, we combine the input image with a random generated image and merge their captions into a single sentence describing both parts of the concatenated image.
In practice we generated a fixed set of 25 images using the base model and always randomly chose one image from this set. We then run the model on the combined image and combined caption, applying the optimized directions only over the tokens from the input prompt. Finally, we apply an $L_2$ loss between the output of the model and the output of the base model only on the part corresponding to the image that was concatenated to the concept image. This encourages the learned directions to only affect the parts in the concept image that match their text.

\ifsiggraph
    \begin{table}
    \centering
    \ifsiggraph
        \small
    ֿ\else
        \small	 
    \fi
    \begin{tabular}{|c|c|c|c|}
    \hline
    
    & Concept & Concept & masks \\
    \textbf{Method} & Decomposition & Composition & free \\
    \hline
    \ifsiggraph
        Dreambooth LoRA & $\times$ & $\times$ & $\surd$ \\ 
    ֿ\else
        DB LoRA & $\times$ & $\times$ & $\surd$ \\ 
    \fi
    Break-A-Scene  & $\surd$ & $\times$ & $\times$ \\ 
    ConceptExpress & $\surd$ & $\times$ & $\surd$\\
    OMG  & $\times$ & $\surd$ & $\surd$ \\ 
    TokenVerse (Ours) & $\surd$  & $\surd$  & $\surd$ \\ 
    \hline
    \end{tabular}
    \caption{\textbf{Capabilities of competing baselines.} The table lists the capabilities of DreamBooth~\cite{ruiz2023dreamboothfinetuningtexttoimage}, BAS~\cite{Avrahami_2023}, ConceptExpress~\cite{hao2024conceptexpressharnessingdiffusionmodels}, OMG~\cite{kong2024omgocclusionfriendlypersonalizedmulticoncept} and our method. 
    \textit{Concept decomposition} is the task of disentangling and personalizing multiple objects from a single image. \textit{Composition} is the task of combining separately learned concepts in a new generated image.
    Unlike existing approaches, our method enables mask-free multi-object disentangled personalization, as well as the composition of multiple objects from several reference images.}
    \label{tab:methods}
\end{table}

    \begin{figure*} [t]
    \centering
    \ifsiggraph
        \includegraphics[width=\linewidth]{images/qual_res_6-1.pdf}
        \vspace{-0.5cm}
    ֿ\else
        \includegraphics[width=\linewidth]{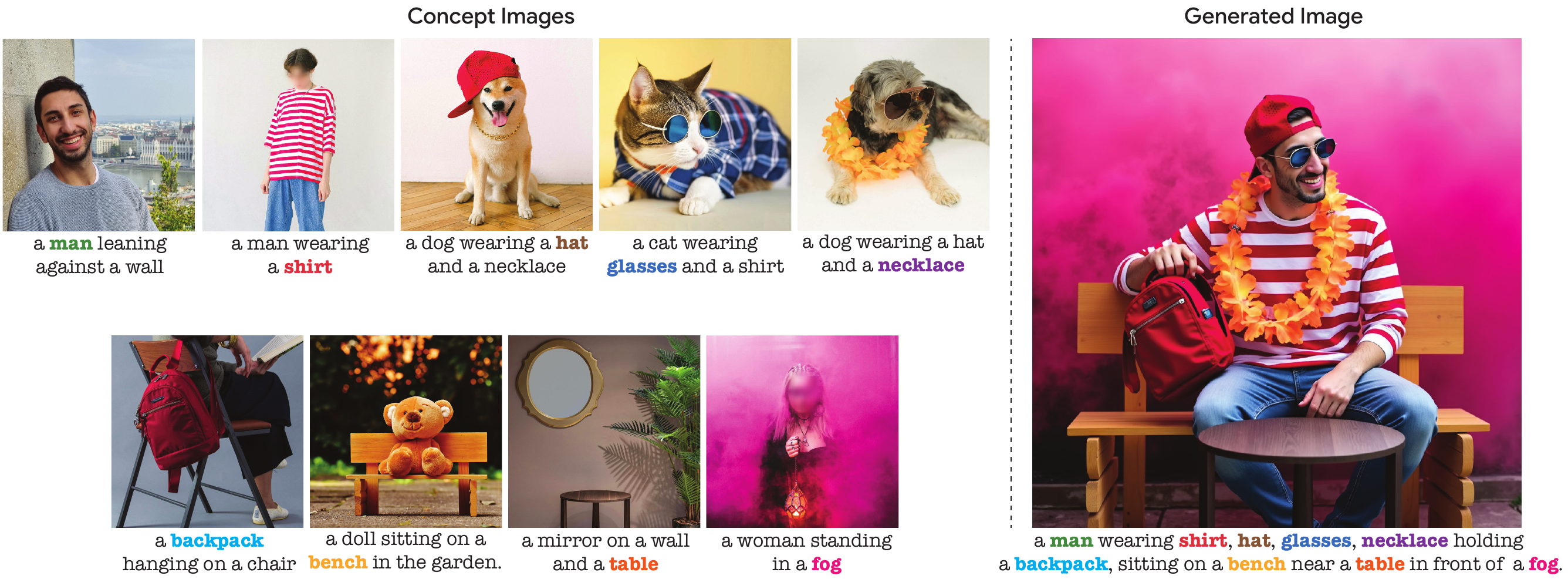}
    \fi
    \caption{\textbf{Extreme multi-concept personalization.} Our method has no technical constraint on the number of concepts that can be combined in an image. As can be seen, TokenVerse can generate images composing a significant number of concepts.
    }
    \label{fig:qualitative_results_6}
    \ifsiggraph
    \vspace{-0.5cm}
    \else
    \fi
\end{figure*}

\fi

\section{Experiments}
We evaluate TokenVerse quantitatively and qualitatively, and demonstrate that it outperforms existing approaches in accurately extracting multiple concepts from images and in seamlessly using them to generate new images.

\subsection{Implementation details}

We illustrate our method with Flux-dev, which has 58 DiT blocks and a modulation vector of dimension 3072. The optimization process involves two stages.  In the first stage, we optimize a global direction for each text token over 800 steps. During this process, for 92\% of the training iterations, we train on timesteps $t$ between 800 and 1000, and for the remaining iterations, we train on timesteps between 0 and 800. In the second stage, we refine the modulation by optimizing a per-block offset for additional 600 steps. In this stage, we reverse the focus by training on $t$ between 0 and 800 for 92\% of the steps and $t$ between 800 and 1000 for the rest. 
We use image augmentations as well as text augmentations. See 
\ifsiggraph
    Supplementary Material (SM)
ֿ\else
    App.~\ref{app:implementation_details}
\fi
for additional details.

For personalization of humans, where capturing precise details is crucial, we use multiple input images depicting the same subject, each accompanied by four augmentations of the input prompt. For all other cases we train on a single concept image with eight augmented prompts.

\subsection{Qualitative results}
\begin{figure}[t]
    \centering
    \ifsiggraph
        \includegraphics[width=\linewidth]{images/pose_light29.pdf}
    ֿ\else
        \includegraphics[width=\linewidth]{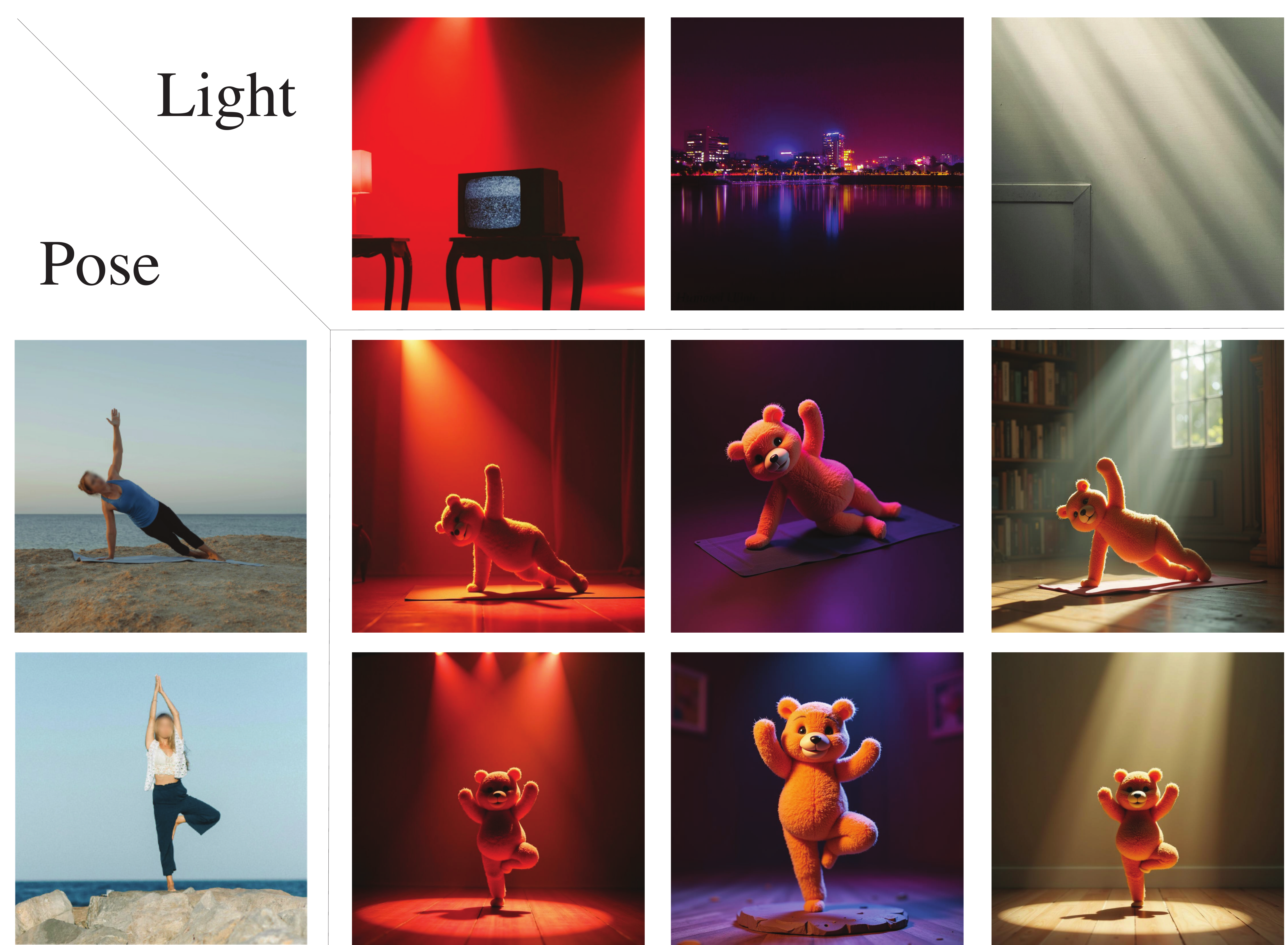}
    \fi
    \caption{\textbf{Concepts beyond objects.} 
    We demonstrate the composition of three types of personalized concepts: object (the bear; concept image not shown), pose (left column) and lighting (top row). 
    TokenVerse successfully learns the pose and lighting without overfitting to the identity of the poser or the specific lit scene.} 
    \label{fig:pose}
\end{figure}

We present qualitative results generated by TokenVerse in 
\ifsiggraph
    Figures~\ref{fig:teaser}, \ref{fig:qualitative_results_6}, \ref{fig:qualitative_results} and \ref{fig:pose}.
ֿ\else
    Figures~\ref{fig:teaser}, \ref{fig:qualitative_results}, \ref{fig:qualitative_results_6} and \ref{fig:pose}.
\fi
Figure~\ref{fig:qualitative_results} presents three generated images for each set of concept images, each depicting a different combination of concepts. 
As can be seen, TokenVerse successfully extracts the desired visual elements from the concept images, and seamlessly combines them in new settings. 
Notably,  TokenVerse correctly disentangles not only the featured objects (\eg the dogs or the cat in the first row) but also their clothing, pose, materials, and even the lighting conditions. 
Importantly, TokenVerse has no technical limitation on the number of concepts that can be combined into a single image.
To illustrate this, Fig.~\ref{fig:qualitative_results_6} showcases the composition of nine personalized objects, each extracted from a different concept image.  
Figure~\ref{fig:pose} further showcases the personalization and combination of abstract concepts such as lighting and pose. As can be seen, the pose concept is completely disentangled from the woman and successfully transferred to the personalized bear. 
More qualitative results, including examples of storytelling, can be found in 
\ifsiggraph
    the SM 
ֿ\else
    App.~\ref{app:qualitative}, App.~\ref{app:storytelling}
\fi
and our project webpage.

\subsection{Comparisons}

\ifsiggraph
\else
    \begin{table}
    \centering
    \ifsiggraph
        \small
    ֿ\else
        \small	 
    \fi
    \begin{tabular}{|c|c|c|c|}
    \hline
    
    & Concept & Concept & masks \\
    \textbf{Method} & Decomposition & Composition & free \\
    \hline
    \ifsiggraph
        Dreambooth LoRA & $\times$ & $\times$ & $\surd$ \\ 
    ֿ\else
        DB LoRA & $\times$ & $\times$ & $\surd$ \\ 
    \fi
    Break-A-Scene  & $\surd$ & $\times$ & $\times$ \\ 
    ConceptExpress & $\surd$ & $\times$ & $\surd$\\
    OMG  & $\times$ & $\surd$ & $\surd$ \\ 
    TokenVerse (Ours) & $\surd$  & $\surd$  & $\surd$ \\ 
    \hline
    \end{tabular}
    \caption{\textbf{Capabilities of competing baselines.} The table lists the capabilities of DreamBooth~\cite{ruiz2023dreamboothfinetuningtexttoimage}, BAS~\cite{Avrahami_2023}, ConceptExpress~\cite{hao2024conceptexpressharnessingdiffusionmodels}, OMG~\cite{kong2024omgocclusionfriendlypersonalizedmulticoncept} and our method. 
    \textit{Concept decomposition} is the task of disentangling and personalizing multiple objects from a single image. \textit{Composition} is the task of combining separately learned concepts in a new generated image.
    Unlike existing approaches, our method enables mask-free multi-object disentangled personalization, as well as the composition of multiple objects from several reference images.}
    \label{tab:methods}
\end{table}

\fi

TokenVerse enables $(i)$ disentangled personalization of multiple concepts from a single image, and $(ii)$ plug-and-play composition of multiple concepts \emph{learned separately} from different images into one generated image. We refer to those sub-tasks as \textit{decomposition} and \textit{composition}, respectively.
Unlike TokenVerse, existing methods address only one of these sub-tasks.
In addition, some of them require object masks as input, which limits their generalization to concepts beyond objects. The capabilities of each method are summarized in Tab.~\ref{tab:methods}.

In order to evaluate these methods on our full task, it is necessary to perform some adaptations, which we describe in  
\ifsiggraph
    the SM.
ֿ\else
    App.~\ref{app:adaptations}.
\fi
Notably, there is no trivial way to adapt Break-A-Scene~\cite{Avrahami_2023} to work on multiple images with separate training, as it requires fine-tuning the model's weights on each concept image. We therefore evaluate it on the easier task of jointly learning the concepts from all concept images together (\ie no plug-and-play composition, but rather a new optimization for each desired combination of objects). For DreamBooth, we provide two adaptations, one for the full task (separate training) and one for the easier task (joint training), as detailed in 
\ifsiggraph
    the SM.
ֿ\else
    App.~\ref{app:adaptations}.
\fi
ConceptExpress, OMG, and TokenVerse are evaluated on the full task.

\subsubsection{Qualitative comparison}
\label{sec:qualitative_comp}
Figure~\ref{fig:qualitative_comp} presents qualitative comparisons on the simplest case of our task: generating an image depicting two concepts, each learned separately from a different concept image. Naturally, each concept image depicts more than one concept. For example, in the first row, the models need to learn the visual attributes of the cat and of the sheep doll, disentangling them from the other elements appearing in the concept images. %
As observed, even in this simplest case, all competing methods fail to disentangle at least one of the two concepts. For instance, in the first row, OMG fails to 
properly identify the doll and extracts the vase instead. This is because it was designed
for personalization from images depicting a single object.

\subsubsection{Quantitative comparison}
\label{sec:quantitative_comp}
We next provide quantitative comparisons on the simplest form of our task, as mentioned above, \ie combining two concepts learned separately from two different images, both containing more than one concept. 
Since some of the baselines are originally designed for either composition or decomposition, we additionally report the performance on each of these sub-tasks separately. 
In the \emph{decomposition experiment}, the goal is to extract two concepts from a single image.
In the \emph{composition experiment}, the goal is to extract a single concept from each of two images that depict only one concept. In both cases, the final goal is to generate an image that combines the two learned concepts.

\ifsiggraph
    \begin{figure*} [t]
    \centering
    \includegraphics[width=\linewidth]{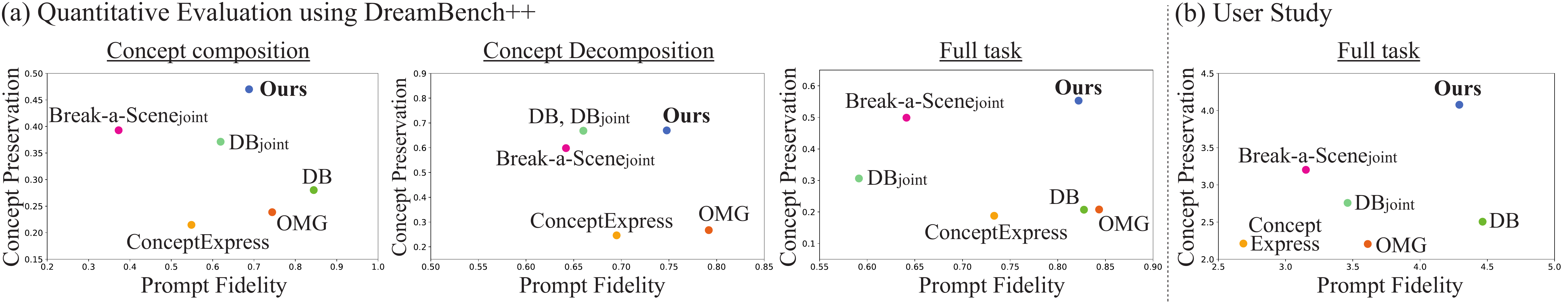}
    \ifsiggraph
    \vspace{-0.75cm}
    \else
    \vspace{-0.5cm}
    \fi

    \caption{\textbf{Quantitative comparison.} 
    We compare our method to other baselines on concept preservation and prompt fidelity (higher is better) using DreamBench++ and a user study. (a) We compare three different settings: $(i)$ composing two concepts from different images (concept composition), $(ii)$ decomposing two concepts from the same image (concept decomposition), and $(iii)$ the combination of the two (full task). (b) We conduct a user study, comparing our method to existing methods on our full task.
    Our method consistently scores best in terms of concept preservation while maintaining high prompt fidelity scores. See 
    \ifsiggraph
        SM
    \else
        App.~\ref{app:quantitative_comparisons}
    \fi
    for the exact metrics.
    }
    \label{fig:quantitative_comparison}
    \ifsiggraph
    \vspace{-0.5cm}
    \else\fi
\end{figure*}

ֿ\else
    \begin{figure*} [t]
    \centering
    \vspace{-0.1cm}
    \includegraphics[width=\linewidth]{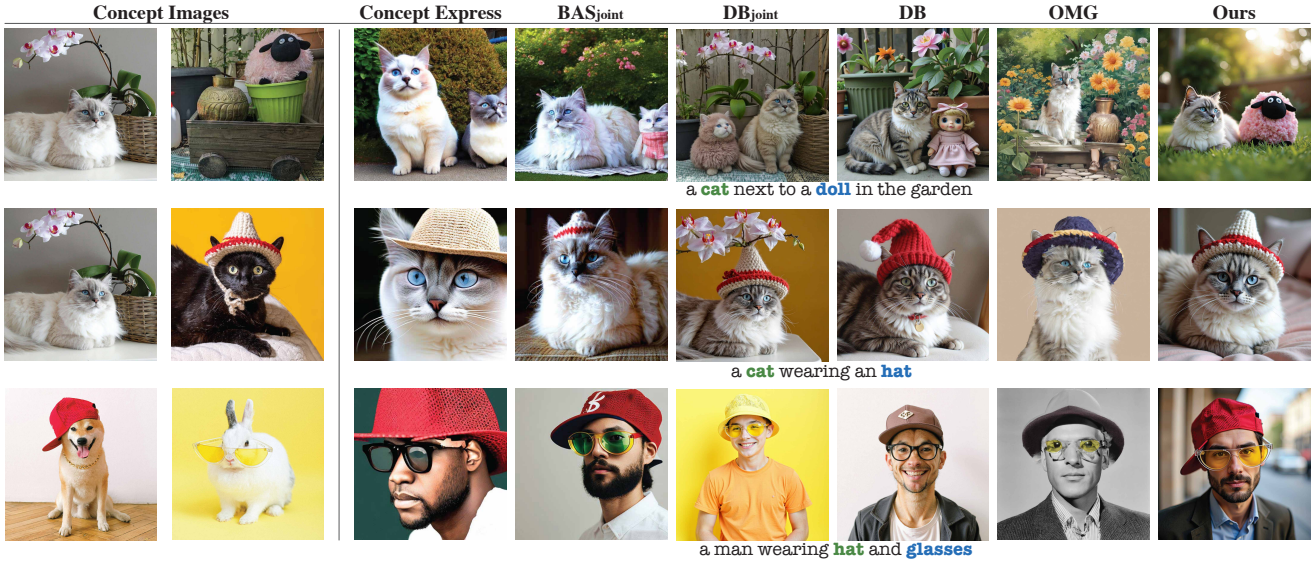}
    \caption{\textbf{Qualitative comparisons.} Each row depicts two concept images (left) and images containing a combination of those concepts, generated by ConceptExpress~\cite{hao2024conceptexpressharnessingdiffusionmodels}, BAS~\cite{Avrahami_2023}, DreamBooth~\cite{ruiz2023dreamboothfinetuningtexttoimage}, OMG~\cite{kong2024omgocclusionfriendlypersonalizedmulticoncept} and our method.  
    The concepts associated with the green and blue words are taken from the left and right concept images, respectively. As can be seen, our method best composes the two concepts while preserving concept fidelity.}
    \label{fig:qualitative_comp}
\end{figure*}

\fi

\paragraph{Evaluation set} %
For images containing a single concept, we used a subset of the dataset provided by DreamBench++ \cite{peng2024dreambenchhumanalignedbenchmarkpersonalized}, which includes 20 images and corresponding target prompts. For concept images that contain multiple concepts, there is no existing benchmark suitable for evaluating our task. We therefore created a custom dataset of 30 images, each containing between two and four distinct concepts. The target prompts for these images were generated using an LLM.

\paragraph{Metrics} 
We follow the evaluation protocol of DreamBench++~\cite{peng2024dreambenchhumanalignedbenchmarkpersonalized}, a benchmark specifically designed to assess personalization in text-to-image models. Given a concept image, a target prompt and the corresponding generated image, DreamBench++ employs a multimodal LLM~\cite{chatgpto} to output two scores in the range of 0 to 1 (higher is better): $(i)$ Concept Preservation (CP) measures how well the personalized concept is preserved in the generated images, and $(ii)$ Prompt Fidelity (PF) assesses how well the generated image adheres to the target prompt. We calculate CP separately for each concept in the generated image, by providing DreamBench++ with a concept image that depicts only the relevant concept where the rest of the image is masked out. 
We report the average CP for the two generated concepts.

\ifsiggraph
ֿ\else
    \begin{figure*} [t]
    \centering
    \includegraphics[width=\linewidth]{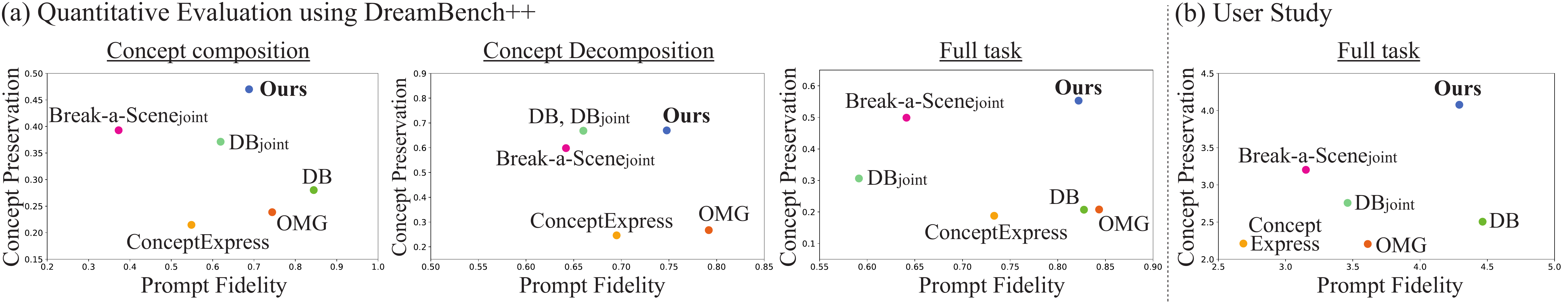}
    \ifsiggraph
    \vspace{-0.75cm}
    \else
    \vspace{-0.5cm}
    \fi

    \caption{\textbf{Quantitative comparison.} 
    We compare our method to other baselines on concept preservation and prompt fidelity (higher is better) using DreamBench++ and a user study. (a) We compare three different settings: $(i)$ composing two concepts from different images (concept composition), $(ii)$ decomposing two concepts from the same image (concept decomposition), and $(iii)$ the combination of the two (full task). (b) We conduct a user study, comparing our method to existing methods on our full task.
    Our method consistently scores best in terms of concept preservation while maintaining high prompt fidelity scores. See 
    \ifsiggraph
        SM
    \else
        App.~\ref{app:quantitative_comparisons}
    \fi
    for the exact metrics.
    }
    \label{fig:quantitative_comparison}
    \ifsiggraph
    \vspace{-0.5cm}
    \else\fi
\end{figure*}

    \begin{figure*}[t]
    \centering
    \includegraphics[width=\linewidth]{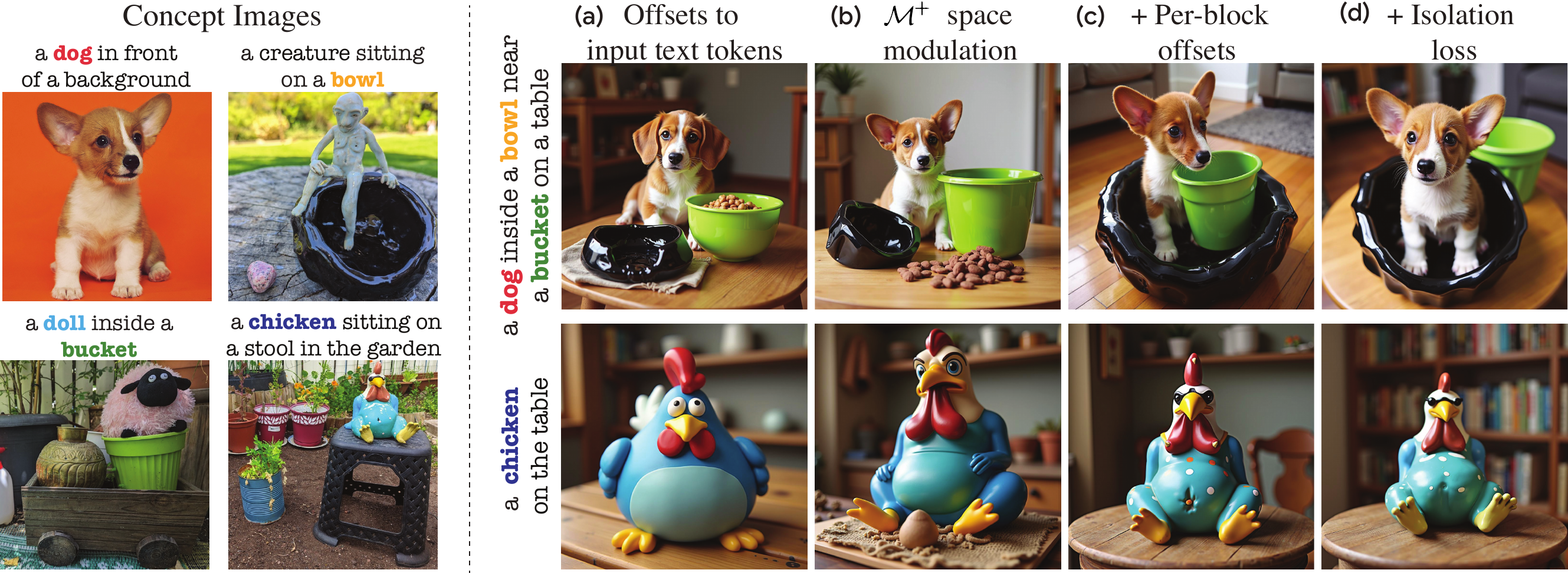}
    \ifsiggraph
    \else
    \vspace{-0.75cm}
    \fi
    \caption{\textbf{Ablations.} The left pane shows all the concepts used to generate the result images. Columns (a) to (d) shows the results of our method as additional components are progressively integrated.}
    \label{fig:ablation_fig}
\end{figure*}

\fi

Figure~\ref{fig:quantitative_comparison}(a-c) quantitatively compares the different methods (see exact numbers in 
\ifsiggraph
    the SM).
ֿ\else
    App.~\ref{app:quantitative_comparisons}).
\fi
As can be seen, TokenVerse consistently outperforms all other methods in terms of Concept Preservation and achieves Prompt Preservation scores that are competitive with the best competitors. %
This applies to three evaluated tasks, demonstrating that $(i)$ TokenVerse outperforms other methods in on the tasks they were originally designed to handle (composition / decomposition) without requiring any visual cues such as segmentation masks; $(ii)$ TokenVerse is superior to other methods on the full task of disentangled concept learning and composition; $(iii)$ TokenVerse outperforms DB$_{\text{joint}}$ and Break-a-Scene$_{\text{joint}}$ even though they were evaluated on the easier task of jointly learning the concepts for each combination. Please refer to 
\ifsiggraph
    the SM
ֿ\else
    App.~\ref{app:quantitative_comparisons}
\fi
for a detailed analysis.

\subsubsection{User study}
We further compare TokenVerse to existing methods by conducting a user study. We select a random subset of examples from the evaluation set described in Sec.~\ref{sec:quantitative_comp}, where each concept image depicts more than one concept and each result image fuses concepts from two different images. 
We followed the user study protocol of Break-A-Scene~\cite{Avrahami_2023} and asked the evaluators to rate both the alignment of the generated image to the prompt and the concept preservation on a scale of 1-5. Since jointly rating the identity preservation of two concepts could cause ambiguity in cases where one concept was preserved well and the other was not, compared to cases where both concepts were slightly preserved, we separated the identity preservation rating into two consecutive questions focusing on one object at a time and averaged the two responses. Our study was conducted with $37$ participants, each
tasked with voting on five results per method, resulting in a total of $3000$ votes. As shown in Fig.~\ref{fig:quantitative_comparison}(d), the user study results are consistent with the DreamBench++ evaluation, giving our method the best concept preservation score with a high prompt fidelity score. Refer to 
\ifsiggraph
    the SM
ֿ\else
    App.~\ref{app:user_study}
\fi
for additional details.

\subsection{Ablation study}
Figure~\ref{fig:ablation_fig} qualitatively illustrates the contribution of each component in TokenVerse.
First, we ablate the use of the $\mathcal{M}^+$ space, by applying the directions predicted by Concept-Mod directly to each text token before it enters the transformer, similarly to textual inversion methods~\cite{gal2022imageworthwordpersonalizing, alaluf2023neuralspacetimerepresentationtexttoimage}.
As can be seen in column (a)
, this approach fails to faithfully reconstruct the concepts. 
In column (b), we show the effect of applying the directions in $\mathcal{M}^+$, \ie adding them to the modulation vector of each text token. As can be seen, this space proves to be more expressive, enabling better concept preservation. Subsequently, we introduce per-block directions in column (c). These further enrich the method’s capabilities and lead to improved concept fidelity. 
However, combining concepts from different images still leads to unsatisfactory results.
Column~(d) shows the results obtained with our full method, including the isolation loss, which mitigates the interference between concepts learned from different images,
and improves concept preservation.

\subsection{Limitations}

While our method is the first to support both the disentangled learning and the composition of multiple visual concepts, some limitations remain for future work. 
First, since the directions are learned separately for each concept image, in rare occasions the resulting modulated text tokens may become similar to each other. In such cases, this similarity could lead to blending of the objects in the generated image, as seen in Fig.~\ref{fig:limitation_fig}(a). We find that this phenomenon happens mostly when the two concepts correspond to the sole or main subjects in their respective images, and is not linked to specific  concept types. In such cases, training on both images jointly fixes the issue. We provide further analysis of this case in 
\ifsiggraph
    the SM
ֿ\else
    App.~\ref{app:limitation_analysis}
\fi
. 
In addition, as seen in Fig.~\ref{fig:limitation_fig}(b), our method struggles with combining two concepts that share the same name identifier. This problem can be mitigated by using distinct words for each concept, as demonstrated in 
\ifsiggraph
    the SM
ֿ\else
    App.~\ref{app:limitation_analysis}
\fi
. Lastly, our method may fail for incompatible combinations. An example is shown in Fig.~\ref{fig:limitation_fig}(c), where attempting to cause a doll with extremely short limbs do a pose that requires arms and legs results in the generation of undesired human body parts.

\ifsiggraph
ֿ\else
    \begin{figure}[t]
    \centering
    \ifsiggraph
     \includegraphics[width=\linewidth]{images/main_limitations.pdf}
ֿ   \else
    \includegraphics[width=\linewidth]{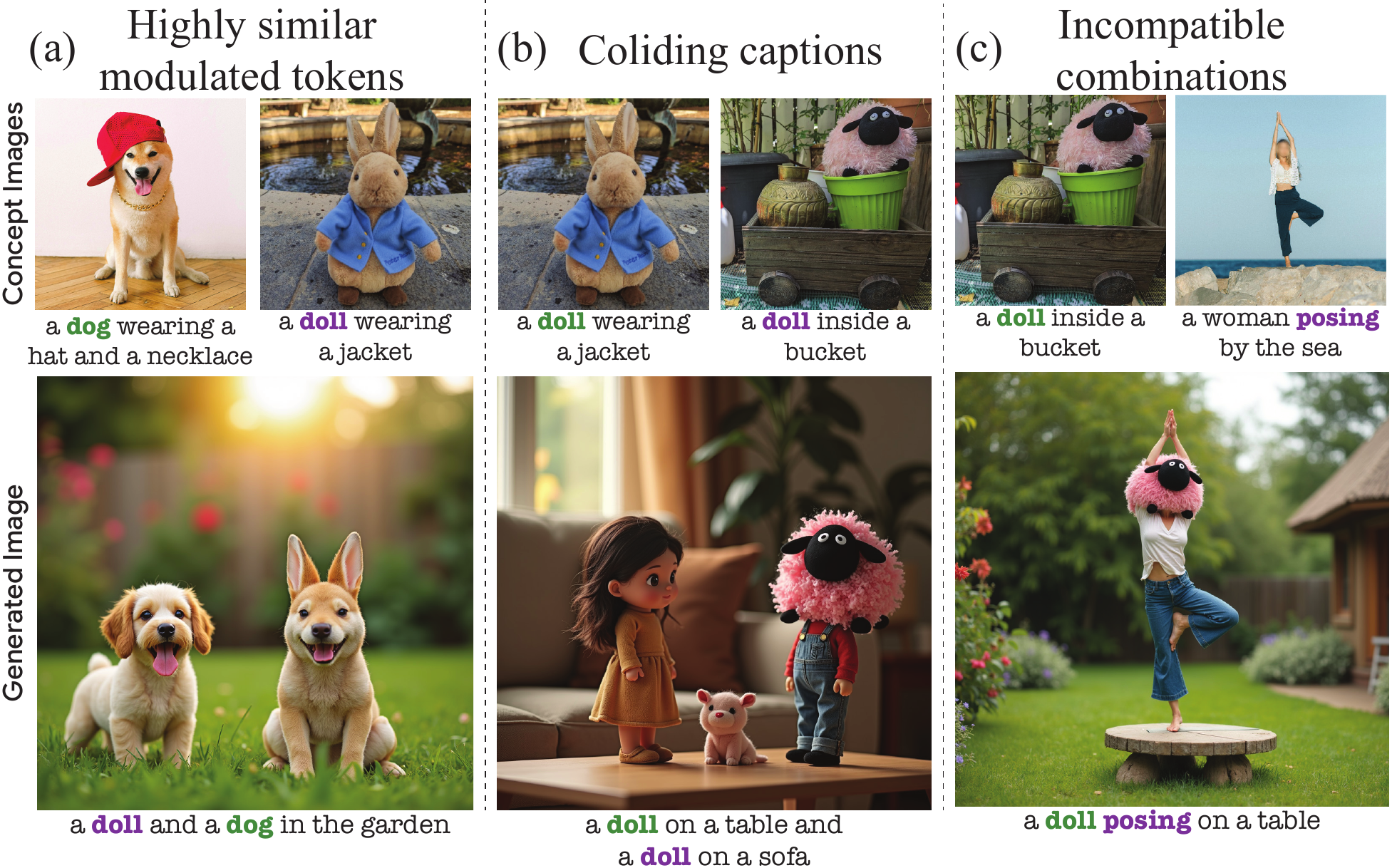}
    \fi
   
    \caption{\textbf{Limitations.}
    Concept images are shown in the top row, with the generated images using TokenVerse below in each case. While our method supports both disentangled learning and multi-concept composition, limitations remain. (a) Rare blending can occur in specific combinations due to independent training of concepts; We provide analysis and mitigations in  
    \ifsiggraph
    the SM.
    ֿ\else
        App.~\ref{app:limitation_analysis}.
    \fi
    (b)~Challenges arise with concepts sharing the same name identifier, which can be mitigated by using distinct terms. (c) Certain incompatible combinations, such as a doll with tiny limbs in a complex pose, may result in undesired outputs.
    } 
    \label{fig:limitation_fig}
\end{figure}

\fi

\section{Conclusion}
Recent progress in text-to-image diffusion models has ignited a surge of interest in harnessing them for controlled image generation. Many approaches have been proposed for personalized image generation, where objects or styles are extracted from reference images and are used to compose new images. However, existing methods struggle to handle multiple images with multiple concepts each, and do not support non-object concepts like poses, materials and lighting conditions. Here, we introduced TokenVerse~--~the first method for multi-concept personalization that overcomes these challenges. TokenVerse extracts per-text-token directions in the modulation space of DiTs, which we showed to be rich and semantic. Our method opens the door to a plethora of applications,  from story telling to personalized content creation.

{
    \small
    \bibliographystyle{ieeenat_fullname}
    \bibliography{main}
}

\ifsiggraph
    \begin{figure*} [t]
    \centering
    \includegraphics[width=\linewidth]{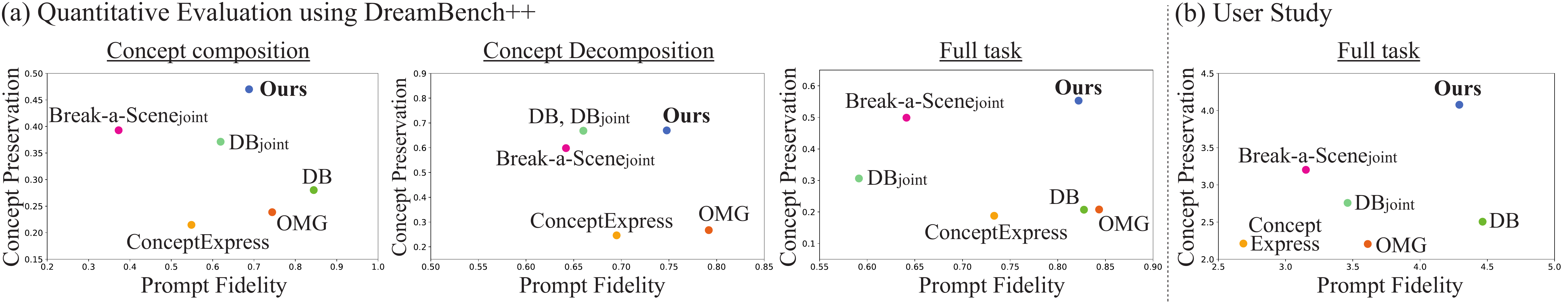}
    \ifsiggraph
    \vspace{-0.75cm}
    \else
    \vspace{-0.5cm}
    \fi

    \caption{\textbf{Quantitative comparison.} 
    We compare our method to other baselines on concept preservation and prompt fidelity (higher is better) using DreamBench++ and a user study. (a) We compare three different settings: $(i)$ composing two concepts from different images (concept composition), $(ii)$ decomposing two concepts from the same image (concept decomposition), and $(iii)$ the combination of the two (full task). (b) We conduct a user study, comparing our method to existing methods on our full task.
    Our method consistently scores best in terms of concept preservation while maintaining high prompt fidelity scores. See 
    \ifsiggraph
        SM
    \else
        App.~\ref{app:quantitative_comparisons}
    \fi
    for the exact metrics.
    }
    \label{fig:quantitative_comparison}
    \ifsiggraph
    \vspace{-0.5cm}
    \else\fi
\end{figure*}

    \appendix
    \maketitlesupplementary
ֿ\else
    \clearpage
    \appendix
    \section*{Appendix}
\fi

\section{Additional training details}
\label{app:implementation_details}

We provide additional training details of our method.

\subsection{Training agmentations}
\label{app:aug}
\begin{figure}[t]
    \centering
    \includegraphics[width=\linewidth]{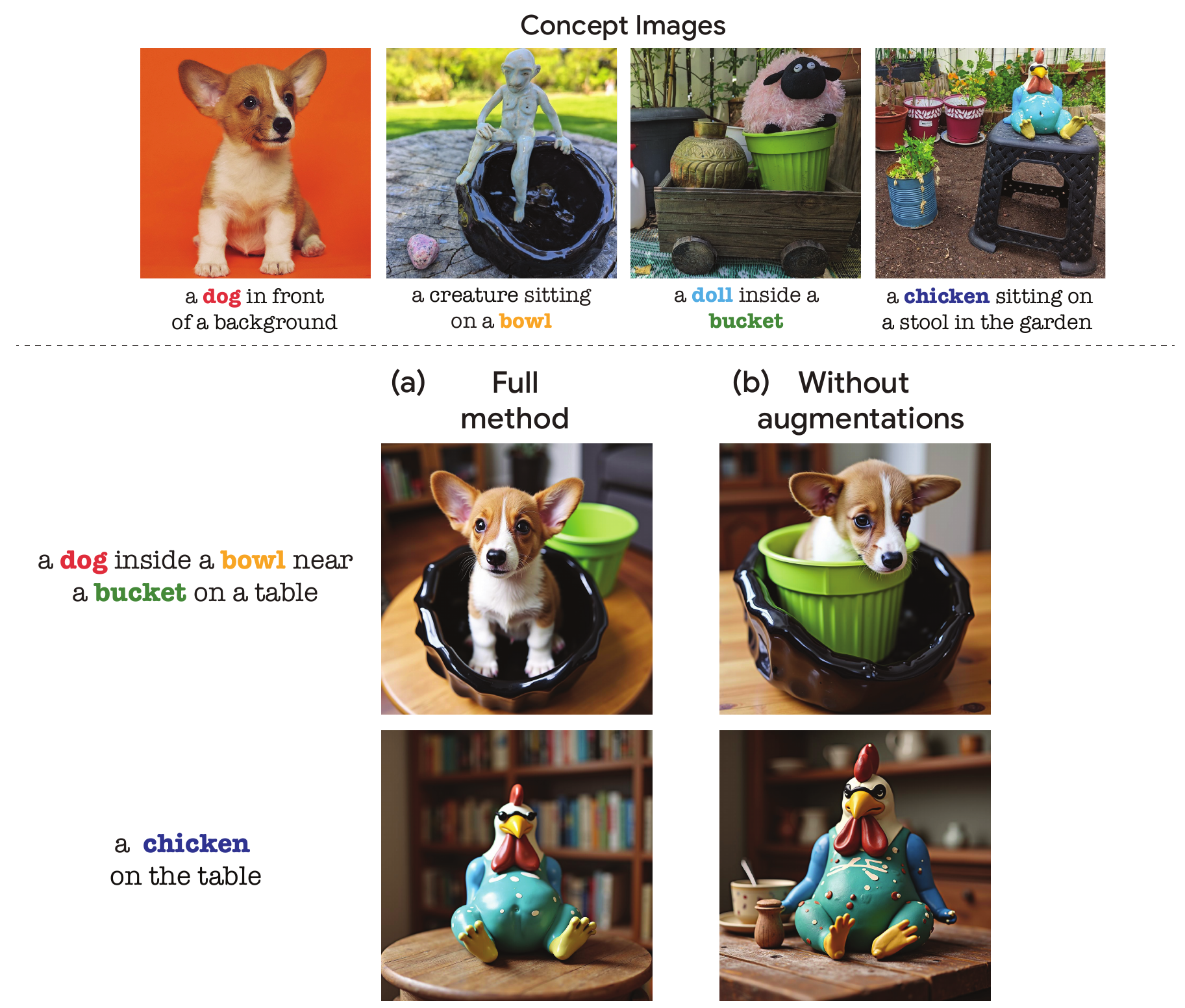}
    \caption{\textbf{Augmentations Ablation.} The top row shows the concepts used to generate the result images. Column (a) displays the results of our full method, while column (b) shows the results without text and image augmentations.}
    \label{fig:ablation_fig_sup}
\end{figure}

During training, each image is paired with a text description.
However, we found empirically that assigning multiple text descriptions per image improves the model's ability to distinguish and separate the different concepts within the image.
We achieve this by augmenting the text with several prompts that closely resemble the original description. Each prompt retains the same words describing the objects and actions but rearranges them in a different order, which can be generated using a large language model (LLM).
In addition to text augmentations, we applied image augmentations such as random flips and mirroring to further enhance the model's performance.
Figure~\ref{fig:ablation_fig_sup} demonstrates the improvement these augmentations have on the resulting images.

\subsection{Images for concept isolation loss}
\label{app:ci_loss}

As described in Sec.~\ref{sec:method} in the main text, our method employs a \textit{concept isolation loss}, where the concept image and its accompanying prompt are concatenated to a random pair of generated image an prompt. In practice, we randomly sample from a fixed set of 25 pairs of captions and generated images for all the examples in the paper. We created these sets of images by asking an LLM to generate 25 prompts that contain objects, and providing the captions to Flux to generate the output images. The entire set of images is presented in Fig.~\ref{fig:il_images}.

\section{Additional qualitative results}
\label{app:qualitative}

\begin{figure}[t]
    \centering
    \ifsiggraph
        \includegraphics[width=\linewidth]{images/plus.pdf}
    ֿ\else
        \includegraphics[width=\linewidth]{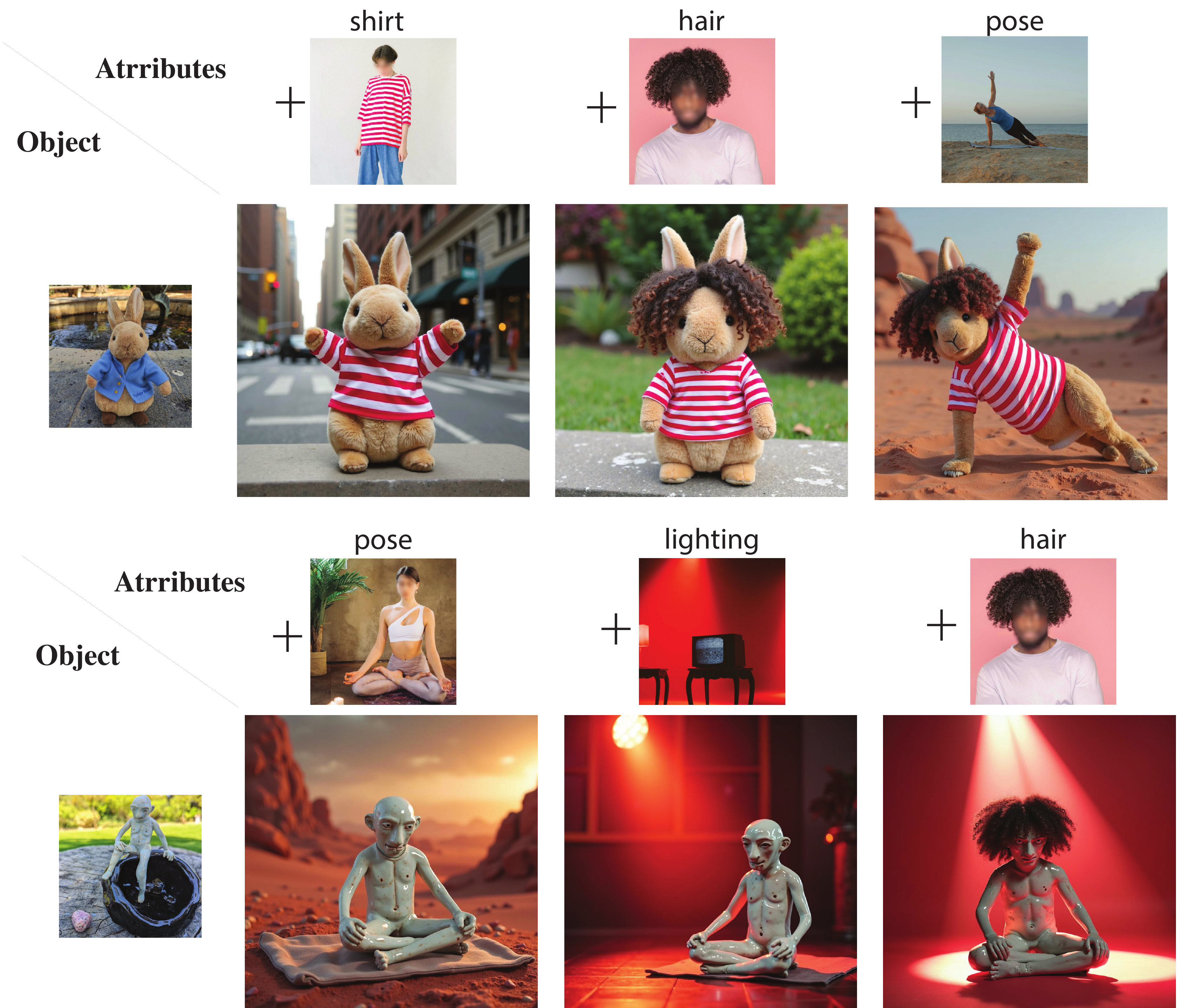}
    \fi
    \caption{\textbf{Progressive composition of concepts.} TokenVerse can be used to progressively add concepts into a generated image, while controlling all other aspects of the generated images via text. In each row, the object, pose, lighting, and hair are personalized, while the background is described by text (\eg ``NY city'', ``garden'', and ``Mars'' for the top row.)}
    \label{fig:compositions}
\end{figure}

\begin{figure*} [t]
    \centering
    \ifsiggraph
        \includegraphics[width=\linewidth]{images/qual_fig_sup_1.pdf}
    ֿ\else
        \includegraphics[width=\linewidth]{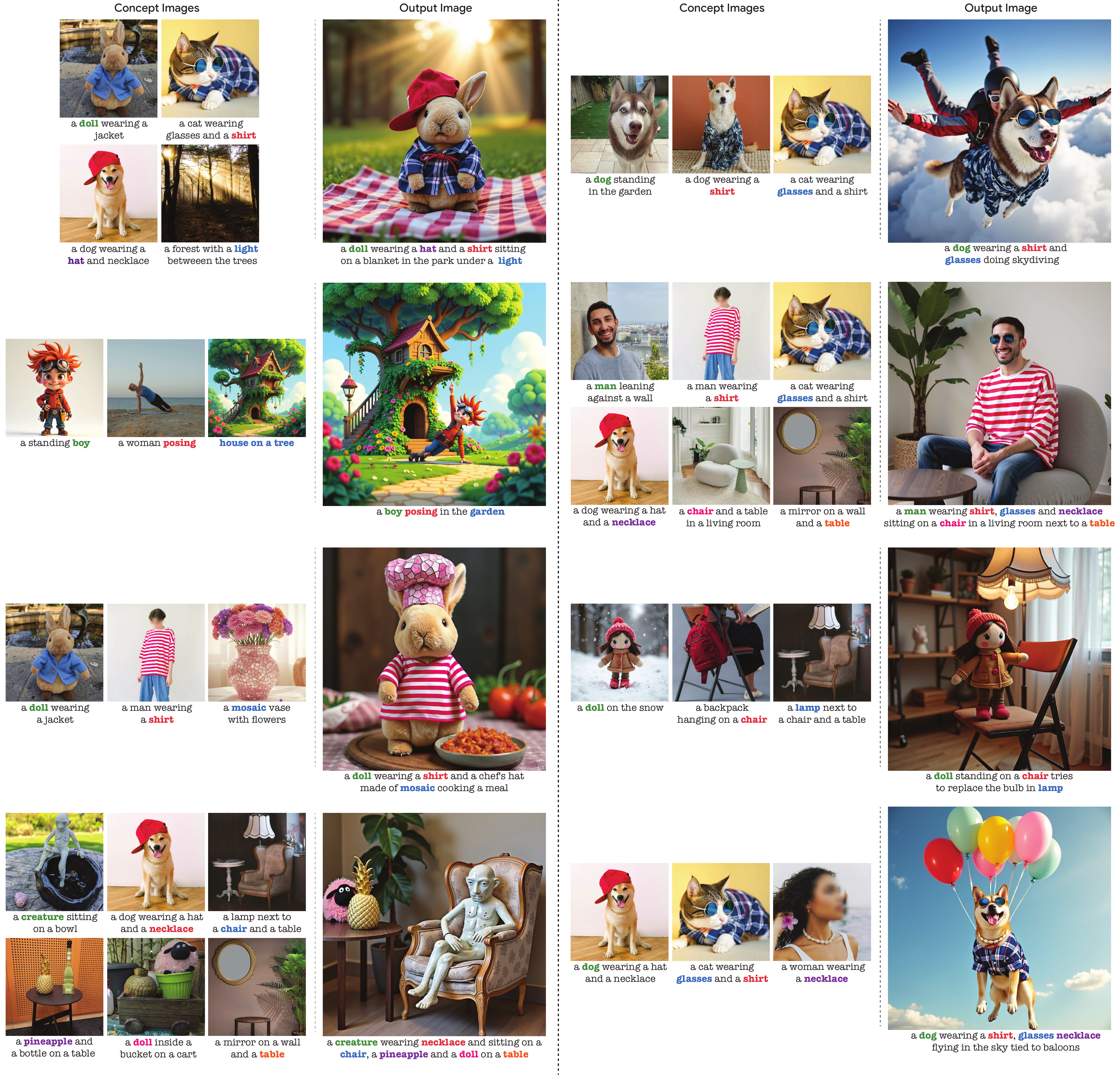}
    \fi
    \vspace{-0.75cm}
    \caption{\textbf{Qualitative results.}
    Each row contains two result images and the source images of the concepts that they contain.
    }
    \label{fig:qualitative_results_sup_1}
\end{figure*}

\begin{figure*} [t]
    \centering
    \ifsiggraph
        \includegraphics[width=\linewidth]{images/qual_fig_sup_2.pdf}
    ֿ\else
        \includegraphics[width=\linewidth]{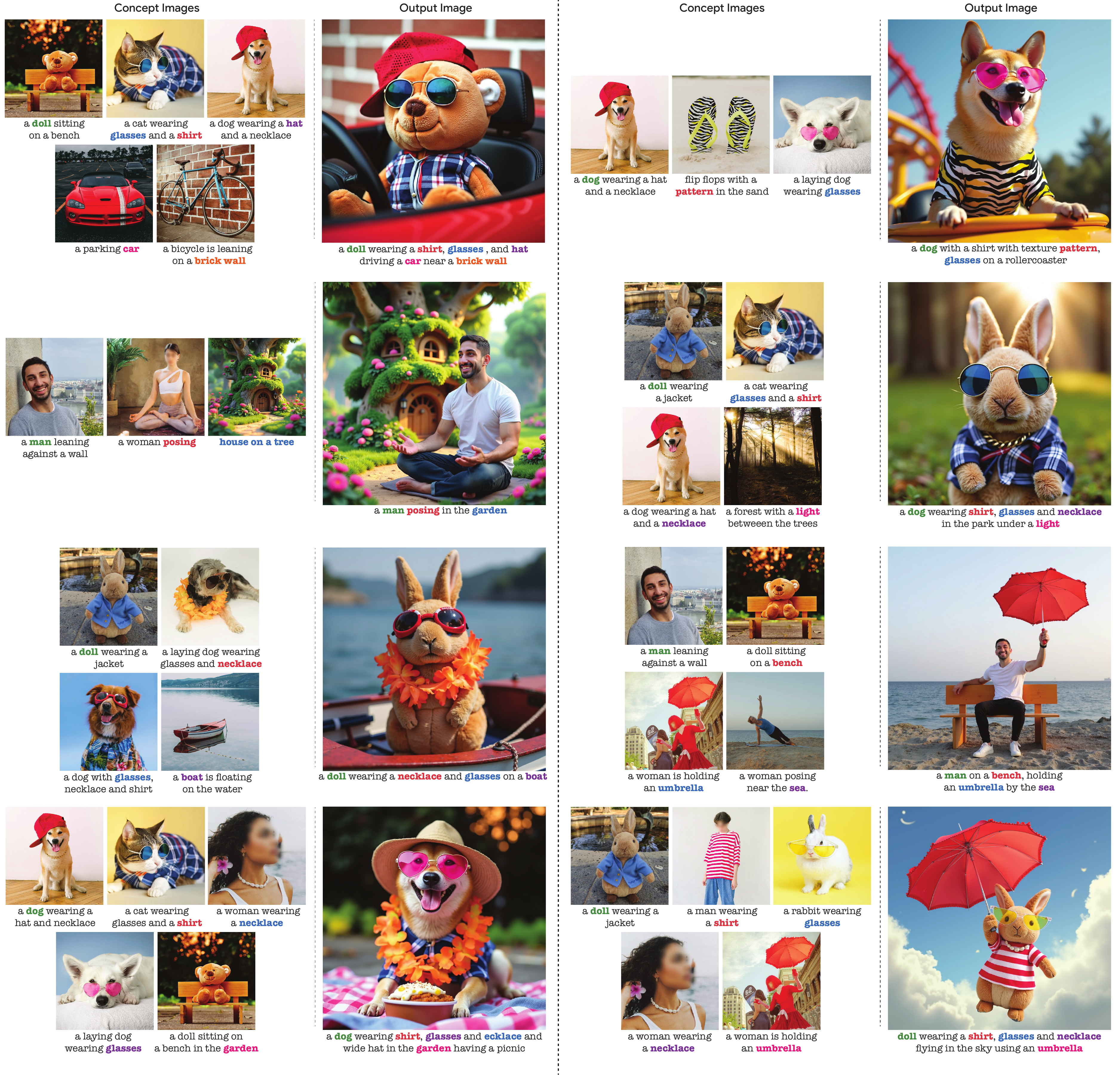}
    \fi
    \vspace{-0.75cm}
    \caption{\textbf{Qualitative results.}
    Each row contains two result images and the source images of the concepts that they contain.
    }
    \label{fig:qualitative_results_sup_2}
\end{figure*}

\begin{figure*} [t]
    \centering
    \ifsiggraph
        \includegraphics[width=\linewidth]{images/qual_fig_sup_3.pdf}
    ֿ\else
        \includegraphics[width=\linewidth]{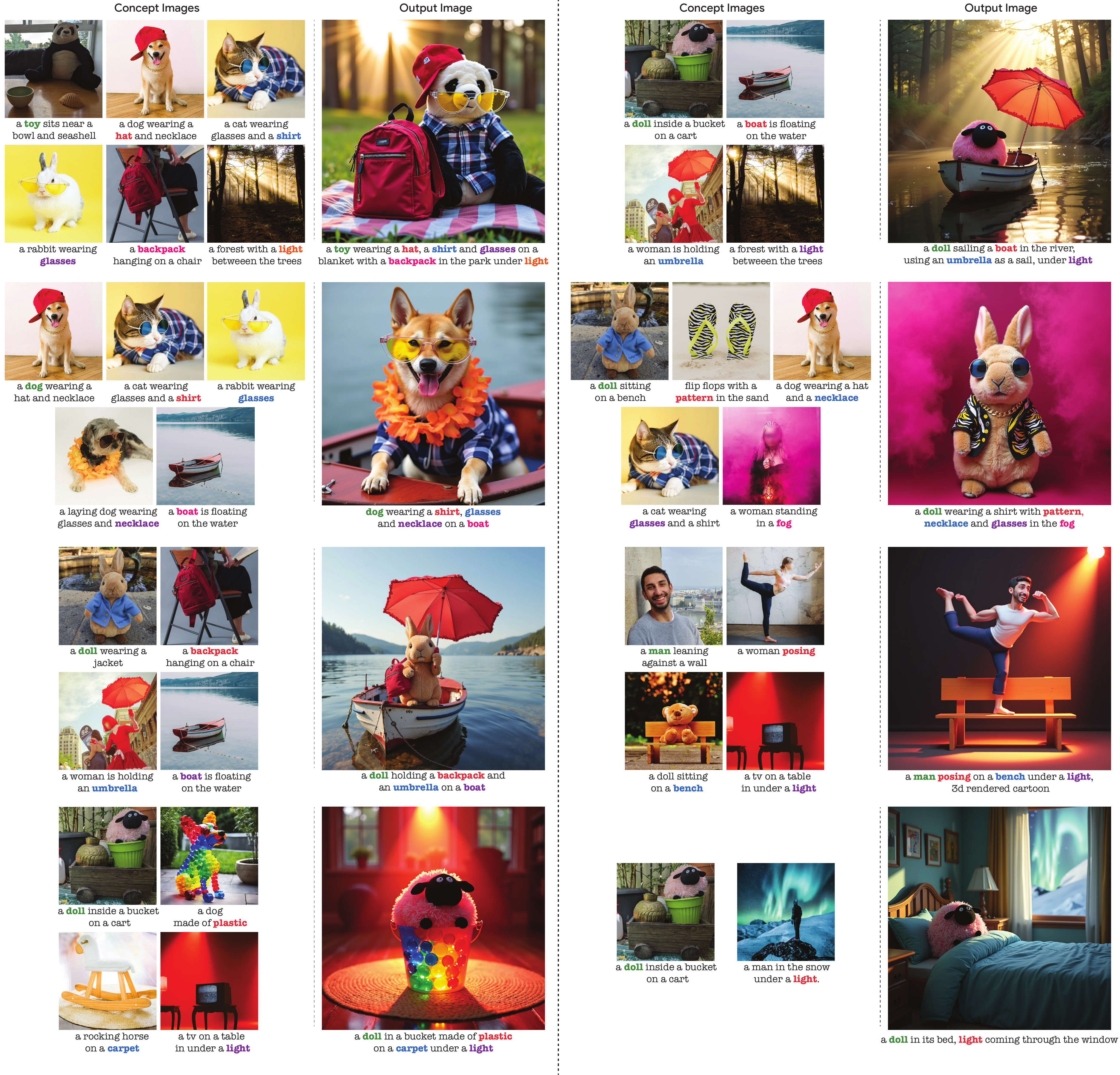}
    \fi
    \vspace{-0.75cm}
    \caption{\textbf{Qualitative results.}
    Each row contains two result images and the source images of the concepts that they contain.
    }
    \label{fig:qualitative_results_sup_3}
\end{figure*}

Figure~\ref{fig:compositions} demonstrates iterative addition of concepts into a generated image. In each column, we introduce an additional learned concept while retaining all previously added concepts. Figures~\ref{fig:qualitative_results_sup_1}, \ref{fig:qualitative_results_sup_2} and \ref{fig:qualitative_results_sup_3}, present additional qualitative examples of our method.

\section{Quantitative evaluation}
\label{app:quantitative_evalaution}

We first describe how we adapted existing methods to our task for evaluation (note that they were also evaluated on their original tasks). Then, we provide the exact metrics from our quantitative evaluation and analyze the results.

\begin{table*}
\centering
\ifsiggraph
ֿ\else
    \small	 
\fi
\begin{tabular}{|c|ccc|ccc|ccc|}
\hline
 & \multicolumn{3}{c|}{\textbf{Composition}} & \multicolumn{3}{c|}{\textbf{Decomposition}} & \multicolumn{3}{c|} {\textbf{Full task}} \\ \hline
\textbf{Metric}                       & CP $\uparrow$ & PF $\uparrow$ & CP $\cdot$ PF $\uparrow$ & CP $\uparrow$ & PF $\uparrow$ & CP $\cdot$ PF $\uparrow$& CP $\uparrow$& PF $\uparrow$& CP $\cdot$ PF$\uparrow$\\  \hline
Dreambooth &  0.280242 & \textbf{0.844422} & \underline{0.236642} & \underline{0.668524} & 0.660167 & \underline{0.441337} & 0.207116 & \underline{0.827521} & 0.171393 \\ 
DreamBooth$_{\text{joint}}$ &  0.371304 & 0.619288 & 0.229944 & \underline{0.668524} & 0.660167 & \underline{0.441337} & 0.306262 & 0.591337 & 0.181104 \\ 
OMG & 0.238606 & \underline{0.743968} & 0.177515 & 0.267477 & 0.\textbf{791793} & 0.211787 & 0.207787 & \textbf{0.843395} & 0.175246 \\ 
Break-A-Scene$_{\text{joint}}$ & \underline{0.392912} & 0.372664 & 0.154380 & 0.598387 & 0.641935 & 0.384126& \underline{0.499054} & 0.641236 & \underline{0.320011} \\ 
ConceptExpress & 0.214718 & 0.548723 & 0.117821 & 0.246387 & 0.695087 & 0.171260 & 0.187853 & 0.733286 & 0.137750 \\ 
\hline
Ours & \textbf{0.470108} & 0.688061 & \textbf{0.323463} & \textbf{0.669940} & \underline{0.747698} &  \textbf{0.500431} & \textbf{0.553125} & 0.821875 & \textbf{0.454600} \\ 
\hline
\end{tabular}
\caption{
\textbf{Dreambench++ Evaluation} We complement the quantitative comparison graphs of the main paper with the exact measurements of concept preservation (CP) and prompt fidelity (PF).
}
\label{tab:quan_tabls}
\end{table*}

\begin{table}
\centering
\begin{tabular}{|c|cc|}
\hline
\textbf{Metric}                       & CP $\uparrow$ & PF $\uparrow$ \\  \hline
Dreambooth & 2.2505 & \textbf{4.465} \\
DreamBooth$_{\text{joint}}$ &  2.7582 & 3.462 \\ 
OMG &  2.205 & 3.611 \\ 
Break-A-Scene$_{\text{joint}}$ &  \underline{3.203} & 3.151 \\ 
ConceptExpress &  2.211 & 2.686 \\ 
\hline
Ours &  \textbf{4.078} & \underline{4.292} \\
\hline
\end{tabular}
\caption{
\textbf{User Study} We complement the user study results graphs of the main paper with the exact measurements of concept preservation and prompt fidelity.
}
\label{tab:user_study_tab}
\end{table}

\subsection{Adaptations to existing methods}
\label{app:adaptations}
ConceptExpress~\cite{hao2024conceptexpressharnessingdiffusionmodels} learns multiple text tokens from a single image without requiring fine-tuning of the model itself, making it directly applicable to our task.
In contrast, OMG~\cite{kong2024omgocclusionfriendlypersonalizedmulticoncept} requires adaptation to support concept decomposition, which we perform by fine-tuning with multiple rare tokens, each linked to a word representing a distinct concept. 
While Break-A-Scene~\cite{Avrahami_2023} is a prominent method for image decomposition, there is no trivial way to adapt it to learn concepts from multiple images with separate training. This is due to the fact that it involves fine-tuning the weights of the model on each concept image. As stated in the main text, we therefore evaluate Break-A-Scene on an easier task, where all of the personalized concepts are learned jointly (\ie performing a separate optimization process per each combination of objects as opposed to the plug-and-play composition supported by our method). We adapt Break-A-Scene to this easier task by fine-tuning the model jointly on all images from which we wish to extract and compose concepts.  We denote the adapted method Break-a-Scene$_{\text{joint}}$.
LoRA-DreamBooth~\cite{ruiz2023dreamboothfinetuningtexttoimage} is the prominent method for personalization (generating one concept learned from one image). We adapt it to concept decomposition similarly to OMG, and compose personalized objects by applying several LoRAs together. We also evaluate an additional variant, DreamBooth$_{\text{joint}}$ on the easier task, by training it jointly on all concept images, as we do for Break-a-Scene$_{\text{joint}}$.

\subsection{Quantitative comparison}
\label{app:quantitative_comparisons}

As described in Sec.~\ref{sec:quantitative_comp}, we follow the evaluation protocol in DreamBench++, which reports Concept Preservation (CP) and Prompt Fidelity (PF).  There often exists a tradeoff between these two criteria. Therefore, in addition to CP and PF, DreamBench++ also reports their product, CP$\cdot$PF, as a unified measure of success.

Table~\ref{tab:quan_tabls} presents the numeric results for the quantitative evaluation described in Sec.~\ref{sec:quantitative_comp}.
As can be seen, our method outperforms all other methods on CP and CP$\cdot$PF across all three tasks. That is, TokenVerse not only surpasses the performance of the adapted methods on our full task, but also outperforms the original (unadapted) methods on the tasks for which they were originally designed to solve. 

Notably, OMG~\cite{kong2024omgocclusionfriendlypersonalizedmulticoncept} achieves very high Prompt Fidelity scores along with an extremely low Concept Preservation scores. We attribute this to the fact that OMG first performs unconditional generation, and then uses a segmentation map obtained from this image as layout for the personalized generation. Therefore, in cases where the personalized object does not match the shape of the corresponding object generated in the first stage, it cannot be accurately incorporated in the target generation. This usually leads to images with sufficient prompt alignment, but which fail to preserve the personalized concepts. This phenomenon is also demonstrated qualitatively in Fig.~\ref{fig:compositions} in the main text, where OMG correctly follows the prompt, yet consistently fails to preserve objects that can vary in shape, such as the sheep doll, the hats and the glasses.
The same type of behavior is also apparent for DreamBooth~\cite{ruiz2023dreamboothfinetuningtexttoimage} whenever the two concepts are learned from different images.
We believe this arises from the combination of two LoRAs, each trained on a separate image, which can compromise the method's ability to preserve the concepts. 
In addition, ConceptExpress~\cite{hao2024conceptexpressharnessingdiffusionmodels} takes an input image and returns a token for each concept it identifies in the image. A manual process is then required to map each concept to its corresponding token. ConceptExpress successfully detects only 75\% of the concepts in the test set. For the remaining concepts, the generation is performed unconditionally.

\section{User study}
\label{app:user_study}
As reported in the main text, we further evaluate the performance of our method by conducting a user study. 
The full numerical results are presented in Table~\ref{tab:user_study_tab}. 

In the study, we randomly sample 15 of the combinations used for the DreamBench++ full task evaluation. This process results in 90 images in total for each rater (6 methods, 15 results each). 

For each image, we asked participants to answer three questions:

\begin{itemize} 
    \item How well does the prompt align with the generated image? 
    \item To what extent does the first concept appear in the generated image? 
    \item To what extent does the second concept appear in the generated image? 
    \end{itemize}

\begin{figure*}[t]
    \centering
    \includegraphics[width=1.0\linewidth]{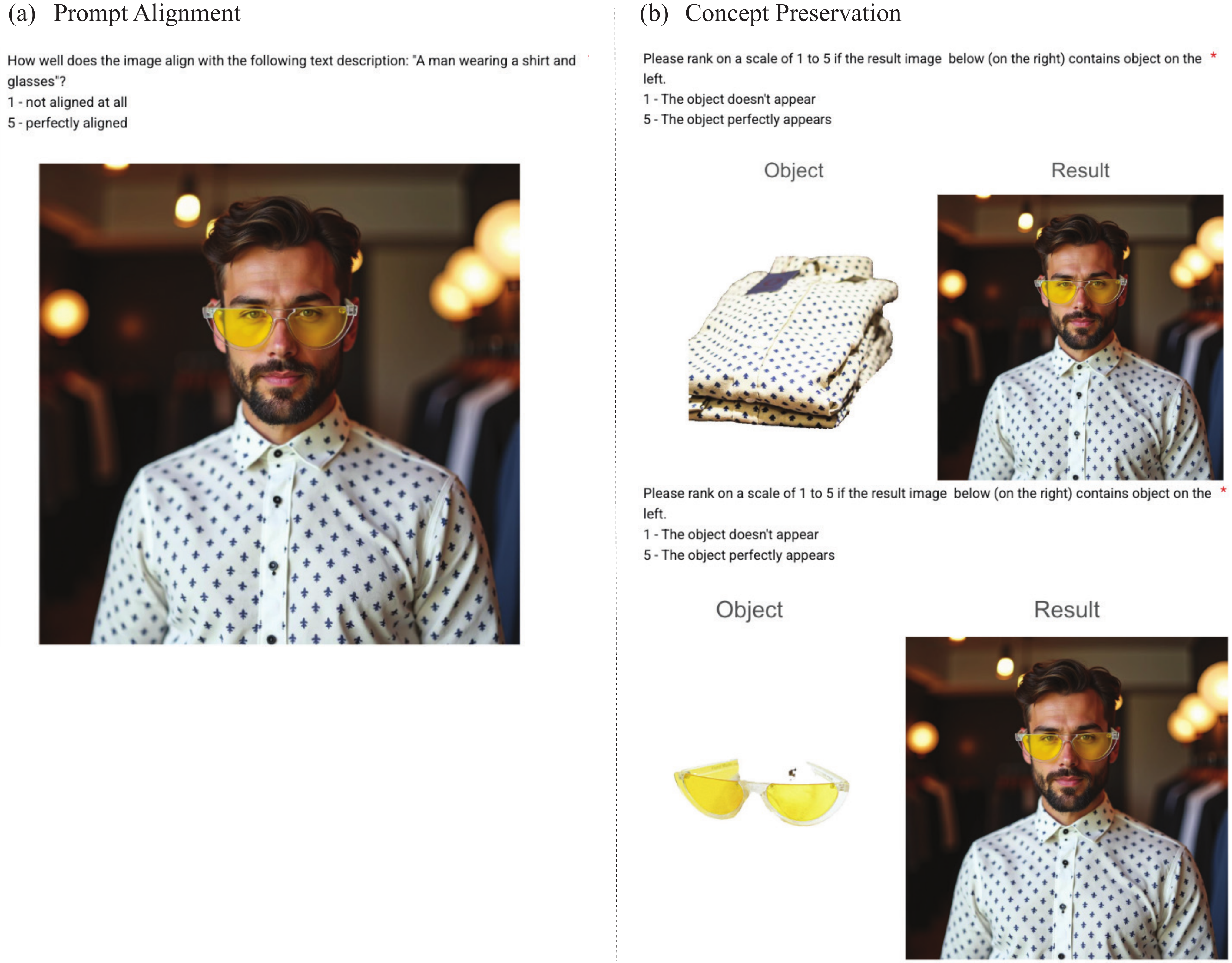}
    \caption{An example of the questions asked in the user study. Given a generated image the users are asked about its alignment with both the text and the input concepts} 
    \label{fig:user_study_example}
\end{figure*}

The exact format of the user study is shown in Fig.~\ref{fig:user_study_example}.

\section{Application to styorytelling}
\label{app:storytelling}
\begin{figure*} [t]
    \centering
    \includegraphics[width=\linewidth]{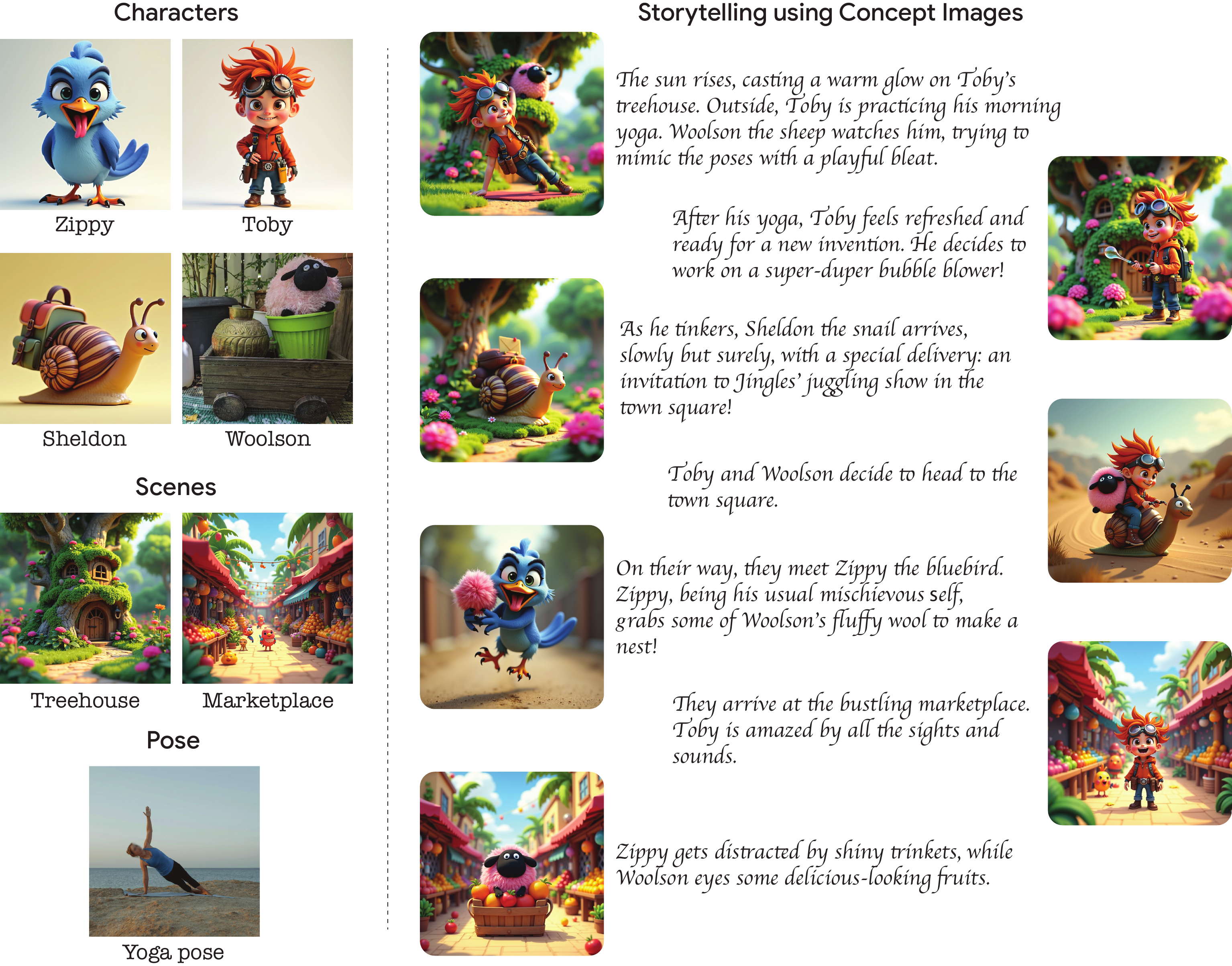}
    \caption{\textbf{Storytelling results.}
    Demonstration of our method's usability for storytelling applications. All the characters, scenes, and poses featured in the story are shown on the left. On the right is the story itself, generated by a language model (LLM). This story was then reprocessed by the LLM to generate prompts, which were used to create the accompanying images.
    }
    \label{fig:storytelling}
\end{figure*}

An immediate application of our method is storytelling, where we aim to generate a narrative consisting of images featuring the same objects and scenes. An example is shown in Fig.~\ref{fig:storytelling}.

\section{Limitation Analysis}
\label{app:limitation_analysis}

\begin{figure}[t]
    \centering
    \includegraphics[width=\linewidth]{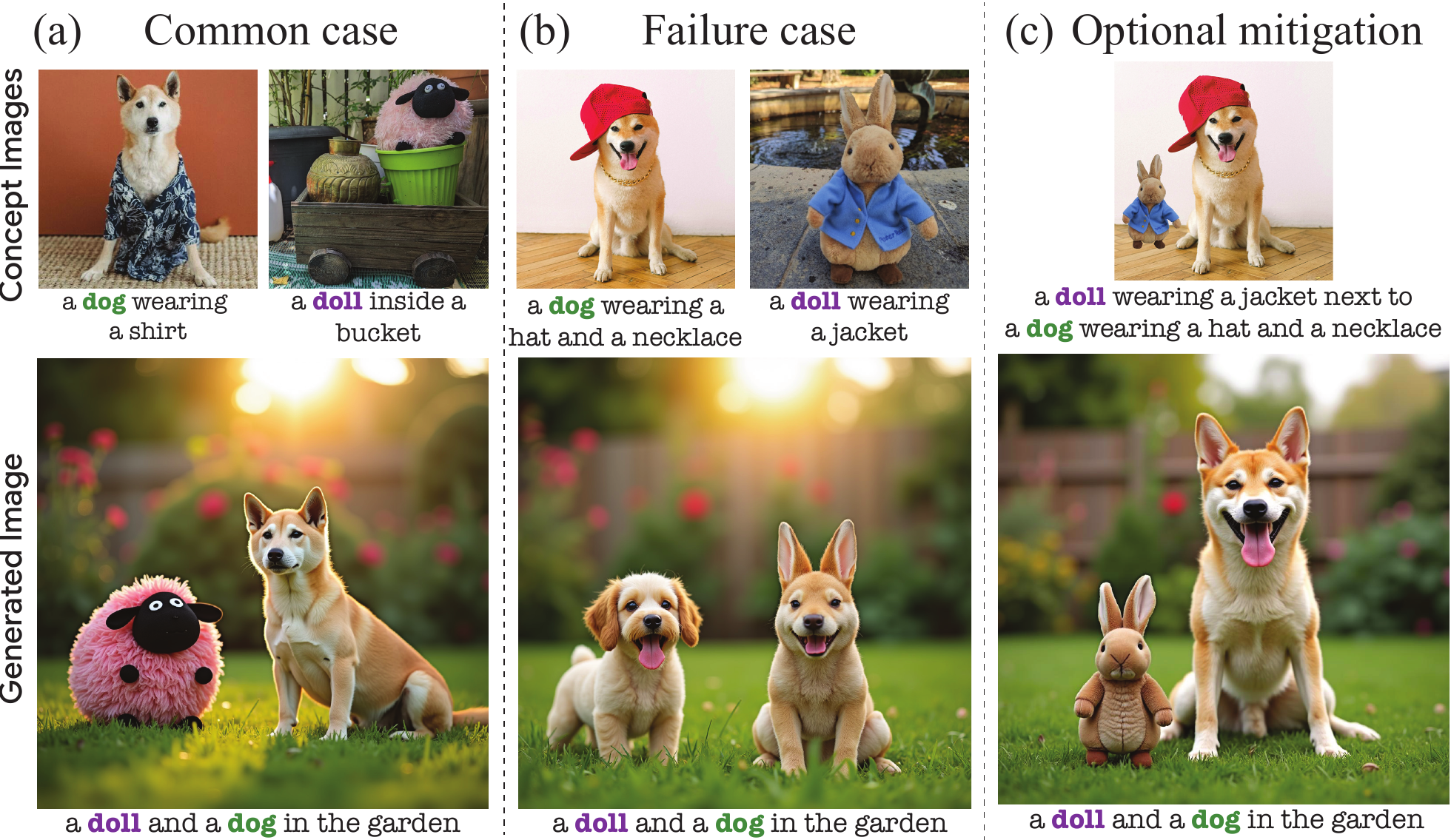}
    \caption{\textbf{Limitation -- highly similar modulated tokens.}
    (a)~The common scenario of combining two distinct objects, such as a doll and a dog, into a single image. (b)~A failure case where independent training of concepts leads to the creation of hybrid objects. (c)~A potential mitigation for this issue by employing joint training on both concepts.
    } 
    \label{fig:limitation_fig_1}
\end{figure}

\begin{figure}[t]
    \centering
    \includegraphics[width=\linewidth]{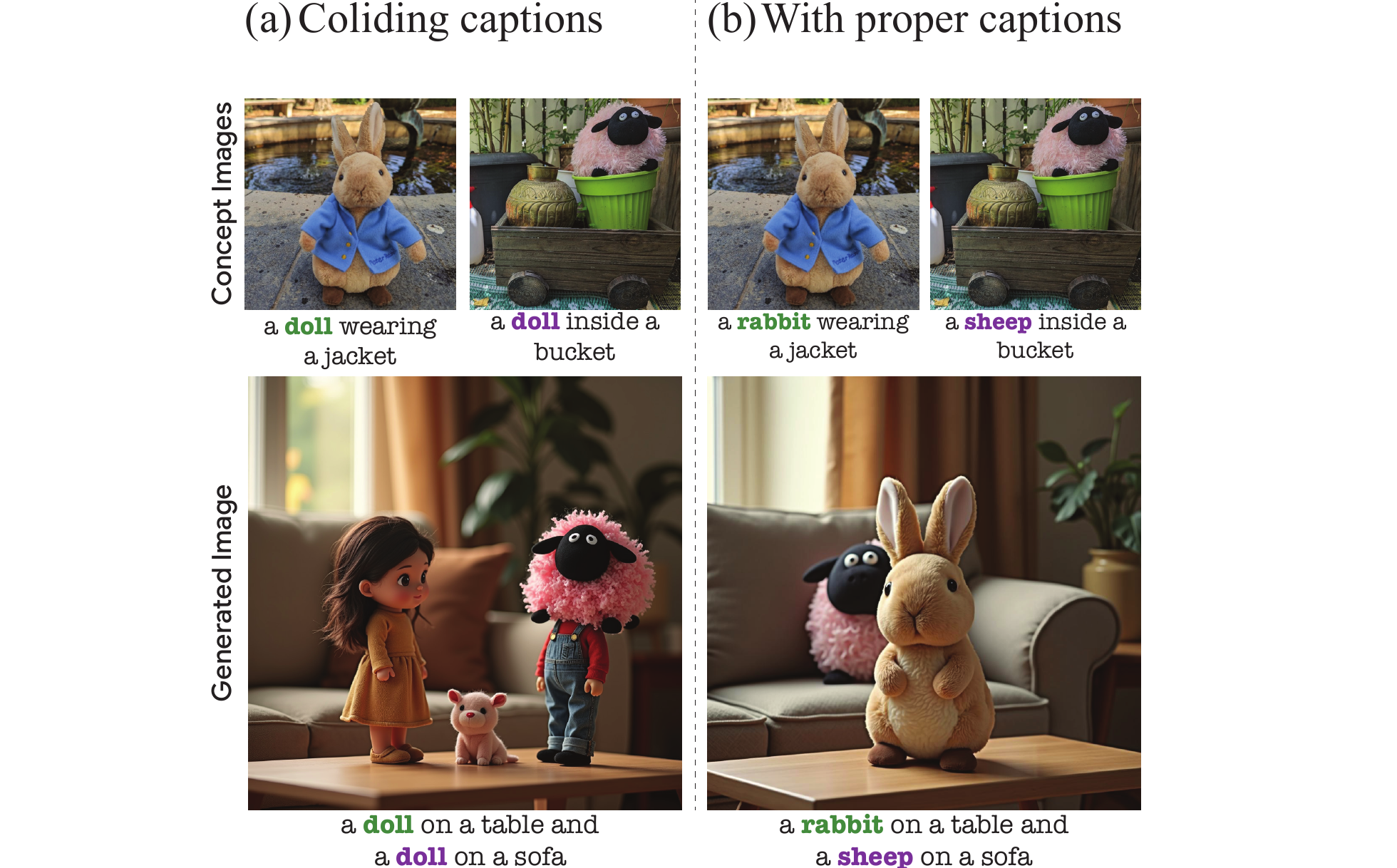}
    \caption{\textbf{Limitations -- colliding captions.} 
    Our method may fail when handling cases of colliding identifiers, such as two dolls (a). This issue can be easily resolved by assigning distinct identifiers to each object during the initial training (b).} 
    \label{fig:limitation_fig_2}
\end{figure}

Figure \ref{fig:limitation_fig_1} shows a possible limitation of our approach. Specifically, in Fig.~\ref{fig:limitation_fig_1}(b) we show that in rare cases, the directions learned for two objects taken from different images may influence each other, leading to undesired results. This behavior does not depend on the type of the objects, as we show in Fig.~\ref{fig:limitation_fig_1}(a) in which TokenVerse successfully generates the concepts ``dog'' and ``doll'' in the same image. 
We observe that in such failure cases, the personalized directions (learned separately for each concept) result in keys with high inner product (at all transformer blocks, for all timesteps), leading to blending in the attention layer. When the directions are learned jointly, the inner product of the keys remains in normal range, and this problem does not occur, as seen in Fig.~\ref{fig:limitation_fig_1}(c). We leave further analysis of this phenomenon to future work.

In addition, as shown in Fig.~\ref{fig:limitation_fig_2}, the choice of word for each token is important. When the same term is used to identify concepts in two different images (\eg ``doll'' in column (a)), the model struggles to generate an image containing both instances. Assigning different identifiers to each object (\eg ``sheep'' for a sheep doll, and ``rabbit'' for the rabbit doll), allows the model to accurately generate an image with both objects as seen in Fig.~\ref{fig:limitation_fig_2}(b).

\begin{figure*} [t]
    \centering
    \includegraphics[width=\linewidth]{images/isolation_loss_small.pdf}
    \caption{\textbf{Generated images used for concept isolation loss.} The images were generated with the base Flux model according to the accompanying prompts.}
    \label{fig:il_images}
\end{figure*}

\end{document}